\documentclass{article}

\usepackage{microtype}
\usepackage{graphicx}
\usepackage{subfigure}
\usepackage{booktabs} 

\usepackage{amsmath}
\usepackage{amsfonts}
\usepackage{bbm}
\usepackage{wrapfig}

\usepackage{amsmath,amsfonts,bm}

\def\eqref#1{equation~\ref{#1}}

\def\1{\bm{1}}

\DeclareMathAlphabet{\mathsfit}{\encodingdefault}{\sfdefault}{m}{sl}
\SetMathAlphabet{\mathsfit}{bold}{\encodingdefault}{\sfdefault}{bx}{n}

\DeclareMathOperator*{\argmax}{arg\,max}

\usepackage[allcolors=blue]{hyperref}

\usepackage[accepted]{icml2021}

\icmltitlerunning{Inverse Constrained Reinforcement Learning}
\begin{document}

\twocolumn[
\icmltitle{Inverse Constrained Reinforcement Learning}



\icmlsetsymbol{equal}{*}

\begin{icmlauthorlist}
\icmlauthor{Usman Anwar}{equal,itu}
\icmlauthor{Shehryar Malik}{equal,itu}
\icmlauthor{Alireza Aghasi}{geo}
\icmlauthor{Ali Ahmed}{itu}
\end{icmlauthorlist}

\icmlaffiliation{itu}{Information Technology University, Lahore, Pakistan}
\icmlaffiliation{geo}{Georgia State University, Atlanta, GA, USA}

\icmlcorrespondingauthor{Usman Anwar}{usman.anwar@itu.edu.pk}

\icmlkeywords{Machine Learning, ICML}

\vskip 0.3in
]



\printAffiliationsAndNotice{\icmlEqualContribution} 

\begin{abstract}
In real world settings, numerous constraints are present which are hard to specify mathematically. However, for the real world deployment of reinforcement learning (RL), it is critical that RL agents are aware of these constraints, so that they can act safely. In this work, we consider the problem of learning constraints from demonstrations of a constraint-abiding agent's behavior. We experimentally validate our approach and show that our framework can successfully learn the most likely constraints that the agent respects. We further show that these learned constraints are \textit{transferable} to new agents that may have different morphologies and/or reward functions. Previous works in this regard have either mainly been restricted to tabular (discrete) settings, specific types of constraints or assume the environment's transition dynamics. In contrast, our framework is able to learn arbitrary \textit{Markovian} constraints in high-dimensions in a completely model-free setting. The code can be found it: \url{https://github.com/shehryar-malik/icrl}.
\end{abstract}

\section{Introduction}

Reward functions are a critical component in reinforcement learning settings. As such, it is important that reward functions are designed accurately and are well-aligned with the intentions of the human designer. This is known as agent (or value) alignment (see, e.g., \citet{leike2018scalable, leike2017gridworlds, amodei2016concrete}). Misspecified rewards can lead to unwanted and unsafe situations (see, e.g, \citet{amodei2016faulty}). However, designing accurate reward functions remains a challenging task. Human designers, for example, tend to prefer simple reward functions that agree well with their intuition and are easily interpretable. For example, a human designer might choose a reward function that encourages an RL agent driving a car to minimize its traveling time to a certain destination. Clearly, such a reward function makes sense in the case of a human driver since inter-human communication is contextualized within a framework of unwritten and unspoken constraints, often colloquially termed as `common-sense'. That is, while a human driver will try to minimize their traveling time, they will be careful not to break traffic rules, take actions that endanger passersby, and so on. However, we cannot assume such behaviors from RL agents since they are are not imbued with common-sense constraints.
\begin{figure}[t]
\vskip 0.2in
\begin{center}
    \subfigure[Expert policy]{\includegraphics[width=0.32\columnwidth]{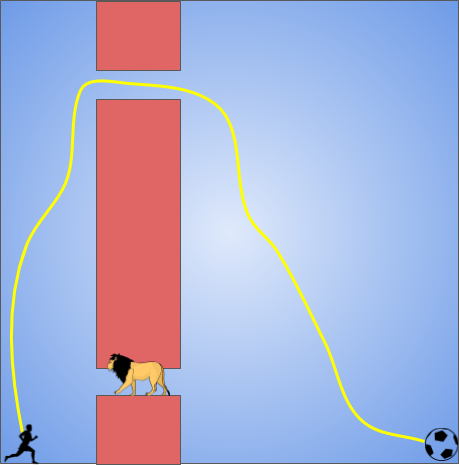}}
    \subfigure[Nominal policy]{\includegraphics[width=0.32\columnwidth]{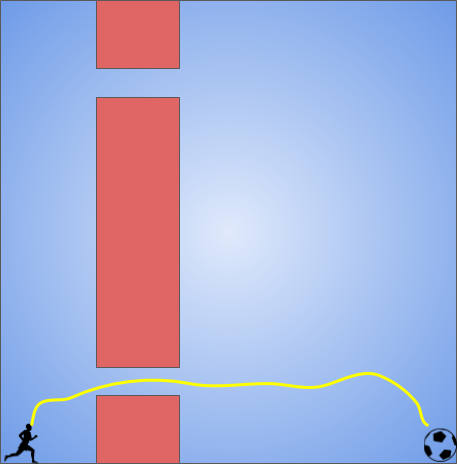}}
    \subfigure[Recovered policy]{\includegraphics[width=0.32\columnwidth]{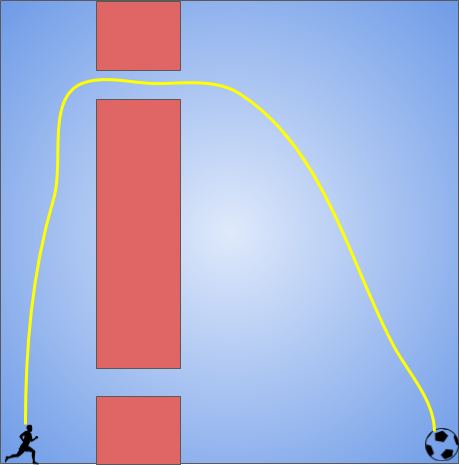}}
    \caption{A simple visualization. The agent's reward is proportional to how close it is to the goal at the bottom-left corner. However, the real world (right) has a lion occupying the lower bridge. The expert is aware of this and so takes the upper bridge. However, the lion is not present in the simulator (middle and left). In a bid to maximize its reward the \textit{nominal} policy chooses to take the lower bridge. Clearly, such a policy will result in the agent being eaten up by the lion in the real world. Our method, on the other hand, tries to reconcile the reward function in the simulator with the expert's behavior, and ultimately learns to avoid the lower bridge.}
    \label{fig:twob}
\end{center}
\vskip -0.2in
\end{figure}

Constrained reinforcement learning provides a natural framework for maximizing a reward function subject to some constraints (we refer the reader to \citet{ray2019safetygym} for a brief overview of the field). However, in many cases, these constraints are hard to specify explicitly in the form of mathematical functions. One way to address this issue is to automatically extract constraints by observing the behavior of a constraint-abiding agent. Consider, for example, the cartoon in Figure \ref{fig:twob}. Agents start at the bottom-left corner and are rewarded according to how quickly they reach the goal at the bottom-right corner. However, what this reward scheme misses out is that in the real world the lower bridge is occupied by a lion which attacks any agents attempting to pass through it. Therefore, agents that are na\"ively trained to maximize the reward function will end up performing poorly in the real world. If, on the other hand, the agent had observed that the expert (in the leftmost figure) actually performed suboptimally with respect to the stipulated reward scheme by taking a longer route to the goal, it could have concluded that (for some unknown reason) the lower bridge must be avoided and consequently would have not been eaten by the lion!

\citet{scobee2020maximum} formalizes this intuition by casting the problem of recovering constraints in the maximum entropy framework for inverse RL (IRL) \citep{ziebart2008maxent} and proposes a greedy algorithm to infer the smallest number of constraints that best explain the expert behavior. However, \citet{scobee2020maximum} has two major limitations: it assumes (1) tabular (discrete) settings, and (2) the environment's transition dynamics. In this work, we aim to address both of these issues by \textit{learning} a constraint function instead through a sample-based approximation of the objective function of \citet{scobee2020maximum}. Consequently, our approach is model-free, admits continuous states and actions and can learn arbitrary Markovian constraints\footnote{Markovian constraints are of the form $c(\tau)=\prod_{t=1}^Tc(s_t,a_t)$, i.e., the constraint function $c$ is independent of the past states and actions in a trajectory. (The notation used here is introduced in the next section.)}. Furthermore, we empirically show that our method scales well to high-dimensions.

Typical inverse RL methods only make use of expert demonstrations and do not assume any knowledge about the reward function at all. However, most reward functions can be expressed in the form ``{do this task while not doing these other things}'' where \textit{other things} are generally constraints that a designer wants to impose on an RL agent. The main task (``do this'') is often quite easy to encode in the form of a simple \textit{nominal} reward function. In this work, we focus on learning the constraint part (``do not do that'') from provided expert demonstrations and using it in conjunction with the nominal reward function to train RL agents. In this perspective, our work can be seen as a principled way to inculcate prior knowledge about the agent's task in IRL. This is a key advantage over other IRL methods which also often end up making assumptions about the agent's task in the form of regularizers such as in \citet{finn2016gcl}.

The main contribution of this work is that it formulates the problem of inferring constraints from a set of expert demonstrations as a \textit{learning} problem which allows it to be used in continuous, high-dimensional settings. To the best of our knowledge, this is the first work in this regard. This also eliminates the need to assume, as \citet{scobee2020maximum} does, the environment's transition dynamics. Finally, we demonstrate the ability of our method to train constraint-abiding agents in high-dimensions and show that it can also be used to prevent reward hacking.

\section{Preliminaries}
\subsection{Unconstrained RL}
A finite-horizon Markov Decision Process (MDP) $\mathcal{M}$ is a tuple $(\mathcal{S}, \mathcal{A}, p, r, \gamma, T)$, where $\mathcal{S} \in \mathbb{R}^{\vert\mathcal{S}\vert}$ is a set of states, $\mathcal{A}\in \mathbb{R}^{\vert\mathcal{A}\vert}$ is a set of actions, $p:\mathcal{S}\times\mathcal{A}\times\mathcal{S} \mapsto [0,1]$ is the transition probability function (where $p(s'\vert s,a)$ denotes the probability of transitioning to state $s'$ from state $s$ by taking action $a$), $r:\mathcal{S} \times \mathcal{A} \mapsto \mathbb{R}$ is the reward function, $\gamma$ is the discount factor and $T$ is the time-horizon. A trajectory $\tau=\{s_1,a_1,\ldots,s_T,a_T\}$ denotes a sequence of states-action pairs such that $s_{t+1} \sim p(\cdot \vert s_t,a_t)$. A policy $\pi : \mathcal{S} \mapsto \mathcal{P}(\mathcal{A})$ is a map from states to probability distributions over actions, with $\pi(a\vert s)$ denoting the probability of taking action $a$ in state $s$. We will sometimes abuse notation to write $\pi(s,a)$ to mean the joint probability of visiting state $s$ and taking action $a$ under the policy $\pi$ and similarly $\pi(\tau)$ to mean the probability of the trajectory $\tau$ under the policy $\pi$.

Define $r(\tau) = \sum_{t=1}^T \gamma^t r(s_t,a_t)$ to be the total discounted reward of a trajectory. Forward RL algorithms try to find an optimal policy $\pi^*$ that maximizes the expected total discounted reward $J(\pi) = \mathbb{E}_{\tau \sim \pi}[r(\tau)]$. On the other hand, given a set of trajectories sampled from the optimal (also referred to as expert) policy $\pi^*$, inverse RL (IRL) algorithms aim to recover the reward function $r$, which can then be used to learn the optimal policy $\pi^*$ via some forward RL algorithm.

\subsection{Constrained RL}
\label{sec:crl}
While normal (unconstrained) RL tries to find a policy that maximizes $J(\pi)$, constrained RL instead focuses on finding a policy that maximizes $J(\pi)$ \textit{while} respecting explicitly-defined constraints. A popular framework in this regard is the one presented in \citet{altman1999cmdps} which introduces the notion of a constrained MDP (CMDP). A CMDP $\mathcal{M}^c$ is a simple MDP augmented with a cost function $c:\mathcal{S}\times\mathcal{A} \mapsto \mathbb{R}$ and a budget $\alpha \geq 0$. Define $c(\tau)=\sum_{t=1}^T \gamma^t c(s_t,a_t)$ to be the total discounted cost of the trajectory $\tau$ and $J^{c}(\pi)=\mathbb{E}_{\tau\sim\pi}[c(\tau)]$ to be the expected total discounted cost. The forward constrained RL problem is to find the policy $\pi_c^*$ that maximizes $J(\pi)$ subject to $J^c(\pi) \leq \alpha$.

In this work, given a set $\mathcal{D}$ of trajectories sampled from $\pi_c^*$, the corresponding unconstrained MDP $\mathcal{M}$ (i.e., $\mathcal{M}^c$ without the cost function $c$) and a budget $\alpha$, we are interested in recovering \textit{a} cost function which when augmented with $\mathcal{M}$ has an optimal policy that generates the same set of trajectories as in $\mathcal{D}$. We call this as the inverse constrained reinforcement learning (ICRL) problem.

If the budget $\alpha$ is strictly greater than $0$, then the cost function $c$ defines \textit{soft} constraints over all possible state-action pairs. In other words, a policy is allowed to visit states and take actions that have non-zero costs as long as the expected total discounted cost remains less than $\alpha$. If, however, $\alpha$ is $0$ then the cost function translates into hard constraints over all state-action pairs that have a non-zero cost associated with them. A policy can thus never visit these state-action pairs. In this work, we restrict ourselves to this hard constraint setting. Note that this is not particularly restrictive since, for example, safety constraints are often hard constraints as well are constraints imposed by physical laws.

Since we restrict ourselves to hard constraints, we can rewrite the constrained RL problems as follows: define $\mathcal{C} = \{(s,a) \vert c(s,a)\neq 0\}$ to be the constraint set induced by $c$. The forward constrained RL problem is to find the optimal constrained policy $\pi^*_\mathcal{C}$ that maximizes $J(\pi)$ subject to $\pi^*_\mathcal{C}(s,a) = 0 \; \forall (s,a) \in \mathcal{C}$. The inverse constrained RL problem is to recover the constraint set $\mathcal{C}$ from trajectories sampled from $\pi^*_\mathcal{C}$.

Finally, we will refer to our unconstrained MDP as the nominal MDP hereinafter. The nominal MDP represents the nominal environment (simulator) in which we train our agent.

\section{Formulation}
\subsection{Maximum Likelihood Constraint Inference}
We take \citeauthor{scobee2020maximum} as our starting point. Suppose that we have a set of trajectories $\mathcal{D}=\{\mathcal{\tau}^{(i)}\}_{i=1}^N$ sampled from an expert $\pi_\mathcal{C}^*$ navigating in a constrained MDP $\mathcal{M}^{\mathcal{C}^*}$ where $\mathcal{C}^*$ denotes the (true) constraint set. Furthermore, we are also given the corresponding nominal MDP $\mathcal{M}$. Our goal is to recover a constraint set which when augmented with $\mathcal{M}$ results in a CMDP that has an optimal policy that respects the same set of constraints as $\pi_\mathcal{C}^*$ does. \citeauthor{scobee2020maximum} pose this as a maximum likelihood problem. That is, if we let $p_\mathcal{M}$ denote probabilities given that we are considering MDP $\mathcal{M}$ and assume a uniform prior on all constraint sets, then we can choose $\mathcal{C}^*$ according to
\begin{equation}
\mathcal{C}^* \leftarrow \argmax_\mathcal{C} p_\mathcal{M}(\mathcal{D}\vert \mathcal{C}).
\label{eq:uniform_prior}
\end{equation}

Under the maximum entropy (MaxEnt) model presented in \citet{ziebart2008maxent}, the probability of a trajectory under a deterministic MDP $\mathcal{M}$ can be modelled as
\begin{equation}
\pi_\mathcal{M}(\tau) = \frac{\exp(\beta r(\tau))}{Z_\mathcal{M}}\mathbbm{1}^\mathcal{M}(\tau),
\label{eq:maxent}
\end{equation}
where $Z_\mathcal{M} = \int \exp(\beta r(\tau))\mathbbm{1}^\mathcal{M}(\tau) d\tau$ is the partition function, $\beta \in [0,\infty)$ is a parameter describing how close the agent is to the optimal distribution (as $\beta \rightarrow \infty$ the agent becomes a perfect optimizer and as $\beta \rightarrow 0$ the agent simply takes random actions) and $\mathbbm{1}$ is an indicator function that is $1$ for trajectories feasible under the MDP $\mathcal{M}$ and $0$ otherwise.

Assume that all trajectories in $\mathcal{D}$ are i.i.d. and sampled from the MaxEnt distribution. We have
\begin{equation}
    p(\mathcal{D} \vert \mathcal{C}) = \frac{1}{\left(Z_{\mathcal{M}^\mathcal{C}}\right)^N}\prod_{i=1}^{N}\exp(\beta r(\tau^{(i)}))\mathbbm{1}^{\mathcal{M}^\mathcal{C}}(\tau^{(i)}).
    \label{eq:p_d_given_c}
\end{equation}
Note that $\mathbbm{1}^{\mathcal{M}^\mathcal{C}}(\tau^{(i)})$ is $0$ for all trajectories that contain any state-action pair that belongs to $\mathcal{C}$. To maximize this, \citeauthor{scobee2020maximum} propose a greedy strategy wherein they start with an empty constraint set and incrementally add state-action pairs that result in the maximal increase in $p(\mathcal{D}\vert\mathcal{C})$.

\subsection{Sample-Based Approximation}
Since we are interested in more realistic settings where the state and action spaces can be continuous, considering all possible state-action pairs individually usually becomes intractable. Instead, we propose a learning-based approach wherein we try to approximate $\mathbbm{1}^{\mathcal{M}^\mathcal{C}}(\tau)$ using a neural network. Consider the log likelihood
\begin{equation}
    \begin{split}
        \mathcal{L(C)}
        &= \frac{1}{N}\log p(\mathcal{D}\vert\mathcal{C})\\
        &= \frac{1}{N}\sum_{i=1}^N\left[\beta r(\tau^{(i)}) + \log \mathbbm{1}^{\mathcal{M}^\mathcal{C}}(\tau^{(i)})\right] - \log Z_{\mathcal{M}^\mathcal{C}}\\
        &= \frac{1}{N}\sum_{i=1}^N\left[\beta r(\tau^{(i)}) + \log \mathbbm{1}^{\mathcal{M}^\mathcal{C}}(\tau^{(i)})\right] - \\&\qquad\qquad\qquad\qquad\log\int\exp(\beta r(\tau))\mathbbm{1}^{\mathcal{M}^\mathcal{C}}(\tau)d\tau.
    \end{split}
\end{equation}
Note that $\mathbbm{1}^{\mathcal{M}^\mathcal{C}}(\tau) = \prod_{t=0}^T \mathbbm{1}^{\mathcal{M}^\mathcal{C}}(s_t, a_t)$ merely tells us whether the trajectory $\tau$ is feasible under the constraint set $\mathcal{C}$ or not. Let us have a binary classifier $\zeta_\theta$ parameterized with $\theta$ do this for us instead, i.e., we want $\zeta_\theta(s_t,a_t)$ to be $1$ if $(s_t, a_t)$ is not in $\mathcal{C}^*$ and $0$ otherwise. Using $\zeta_\theta(\tau)$ as a short hand for $\prod_{t=0}^T \zeta_\theta(s_t, a_t)$, we have 
\begin{equation}
    \begin{split}
        \mathcal{L}(\mathcal{C})
        &= \mathcal{L}(\theta)
        = \frac{1}{N}\sum_{i=1}^N\left[\beta r(\tau^{(i)}) + \log \zeta_\theta(\tau^{(i)})\right] \\&\qquad\qquad\qquad\qquad-\log\int\exp(\beta r(\tau))\zeta_\theta(\tau)d\tau.
    \end{split}
	\label{eq:ll}
\end{equation}
Let $\mathcal{M}^{\bar{\zeta}_\theta}$ denote the MDP obtained after augmenting $\mathcal{M}$ with the cost function $\bar{\zeta}_\theta := 1-\zeta_\theta$\footnote{Note that since we are assuming that $\alpha$ is $0$, we can assign any non-zero (positive) cost to state-action pairs that we want to constrain. Here $1-\zeta_\theta$ assigns a cost of $1$ to all such pairs.}, and $\pi_{\mathcal{M}^{\bar{\zeta}_\theta}}$ denote the corresponding MaxEnt policy. Taking gradients of (\ref{eq:ll}) with respect to $\theta$ gives us (see Section \ref*{a:gll} in the supplementary material for derivation)
\begin{equation}
        \nabla_\theta\mathcal{L}(\theta) = \frac{1}{N}\sum_{i=1}^N\nabla_\theta \log \zeta_\theta(\tau^{(i)}) - \mathbb{E}_{\tau \sim \pi_{\mathcal{M}^{\bar{\zeta}_\theta}}}\left[\nabla_\theta \log \zeta_\theta(\tau)\right].
    \label{eq:grad_theta_exact}
\end{equation}
Using a sample-based approximation for the right-hand term we can rewrite the gradient as
\begin{equation}
   \nabla_\theta\mathcal{L}(\theta) \approx \frac{1}{N}\sum_{i=1}^N\nabla_\theta \log \zeta_\theta(\tau^{(i)}) - \frac{1}{M}\sum_{j=1}^M\nabla_\theta \log \zeta_\theta(\hat{\tau}^{(j)}),
   \label{eq:grad_theta}
\end{equation}
where $\hat{\tau}$ are sampled from $\pi_{\mathcal{M}^{\bar{\zeta}_\theta}}$ (discussed in the next section). Notice that making $\nabla_\theta \mathcal{L}(\theta)$ zero essentially requires matching the expected gradient of $\log \zeta_\theta$ under the expert (left hand term) and nominal (right hand term) trajectories. For brevity, we will write $\pi_{\mathcal{M}^{\bar{\zeta}_\theta}}$ as $\pi_{\theta}$ from now onwards. We can choose $\zeta_\theta$ to be a neural network with parameters $\theta$ and a sigmoid at the output. We train our neural network via gradient descent by using the expression for the gradient given above.

In practice, since we have a limited amount of data, $\zeta_\theta$, parameterized as a neural network, will tend to overfit. To mitigate this, we incorporate the following regularizer into our objective function.
\begin{equation}
    R(\theta) = - \delta\sum_{\tau \sim \{\mathcal{D},\mathcal{S}\}} \left\vert 1 - \zeta_{\theta}(\tau) \right \vert
    \label{eq:reg}
\end{equation}
where $\mathcal{S}$ denotes the set of trajectories sampled from $\pi_\theta$ and $\delta \in [0,1)$ is a fixed constant. $R$ incentivizes $\zeta_\theta$ to predict values close to $1$ (recall that $\zeta_\theta\in(0,1)$), thus encouraging it to choose the smallest number of constraints that best explain the expert data.

\subsection{Forward Step}
\label{sec:fs}
To evaluate (\ref{eq:grad_theta}) we need to sample from $\pi_{\theta}$. Recall that $\pi_{\theta}$ needs to maximize $J(\pi)$ subject to $\pi_{\theta}(s,a) = 0\; \forall (s,a) \in \mathcal{Z}$ where $\mathcal{Z}$ is the constraint set induced by $\bar{\zeta}_\theta$. However, since $\zeta_\theta$ outputs continuous values in the range $(0,1)$, we instead solve the \textit{soft} constrained RL version, wherein we try to find a policy $\pi$ that maximizes $J(\pi)$ subject to $\mathbb{E}_{\tau\sim\pi}[\bar{\zeta}_\theta(\tau)] \leq \alpha$. In our experiments, we set $\alpha$ to a very small value. Note that if $\alpha$ is strictly set to $0$ our optimization program will have an empty feasible set.

We represent our policy as a neural network with parameters $\phi$ and train it through the constrained policy optimization algorithm introduced in \citet{tessler2018reward}, which essentially tries to solves the following equivalent unconstrained min-max problem on the Lagrangian of the objective function
\begin{equation}
	\min_{\lambda \geq 0} \max_\phi J(\pi^{\phi}) +\frac{1}{\beta}\mathcal{H}(\pi^{\phi}) - \lambda (\mathbb{E}_{\tau\sim\pi^{\phi}}[\bar{\zeta}_\theta(\tau)] - \alpha)
	\label{eq:min_max}
\end{equation}
by gradient ascent on $\phi$ (via the Proximal Policy Optimization (PPO) algorithm \citep{schulman2017proximal}) and gradient descent on the Lagrange multiplier $\lambda$. Note that we also add the entropy $\mathcal{H}(\pi^{\phi}) = -\mathbb{E}_{\tau\sim\pi^\phi}[\log \pi^\phi(\tau)]$ of $\pi^\phi$ to our objective function. Maximizing the entropy ensures that we recover the MaxEnt distribution given in (\ref{eq:maxent}) at convergence (see Section \ref*{a:maxent} in the supplementary material for proof).

\subsection{Incorporating Importance Sampling}
Running the forward step in each iteration is typically very time consuming. To mitigate this problem, instead of approximating the expectation in (\ref{eq:grad_theta_exact}) with samples from $\pi_{\theta}$, we approximate it with samples from an older policy $\pi_{\bar{\theta}}$ where $\bar{\theta}$ were the parameters of $\zeta$ at some previous iteration. We, therefore, only need to learn a new policy after a fixed number of iterations. To correct for the bias introduced in the approximation because of using a different sampling distribution, we add importance sampling weights to our expression for the gradient. In this case, the per step importance sampling weights can be shown to be given by (see Section \ref*{a:is} in the supplementary material for derivation)
\begin{equation}
	\omega(s_t,a_t) = \frac{\zeta_\theta(s_t,a_t)}{\zeta_{\bar{\theta}}(s_t,a_t)}.
	\label{eq:is_weights}
\end{equation}
Note that the importance sampling weights at each timestep only depend on the state and action at that timestep (and not the past and/or future states and actions). The gradient can thus be rewritten as
\begin{equation}
    \begin{split}
	    \nabla_\theta\mathcal{L}(\theta) &\approx \frac{1}{N}\sum_{i=1}^N\sum_{s_t,a_t\in\tau^{(i)}} \nabla_\theta\log \zeta_\theta(s_t,a_t) -\\&\qquad\frac{1}{M}\sum_{j=1}^M \sum_{\hat{s}_t,\hat{a}_t\in\tau^{(j)}} \omega(\hat{s}_t,\hat{a}_t) \nabla_\theta \log \zeta_\theta(\hat{s}_t,\hat{a}_t),
	\label{eq:grad_theta_is}
	\end{split}
\end{equation}
where $\{\tau^{(j)}\}_{j=1}^M$ are sampled from $\pi_{\bar\theta}$.

\begin{figure*}[ht!]
\vskip 0.2in
\begin{center}
    \subfigure[LapGridWorld]{\label{fig:lapgw}\includegraphics[width=0.19\textwidth]{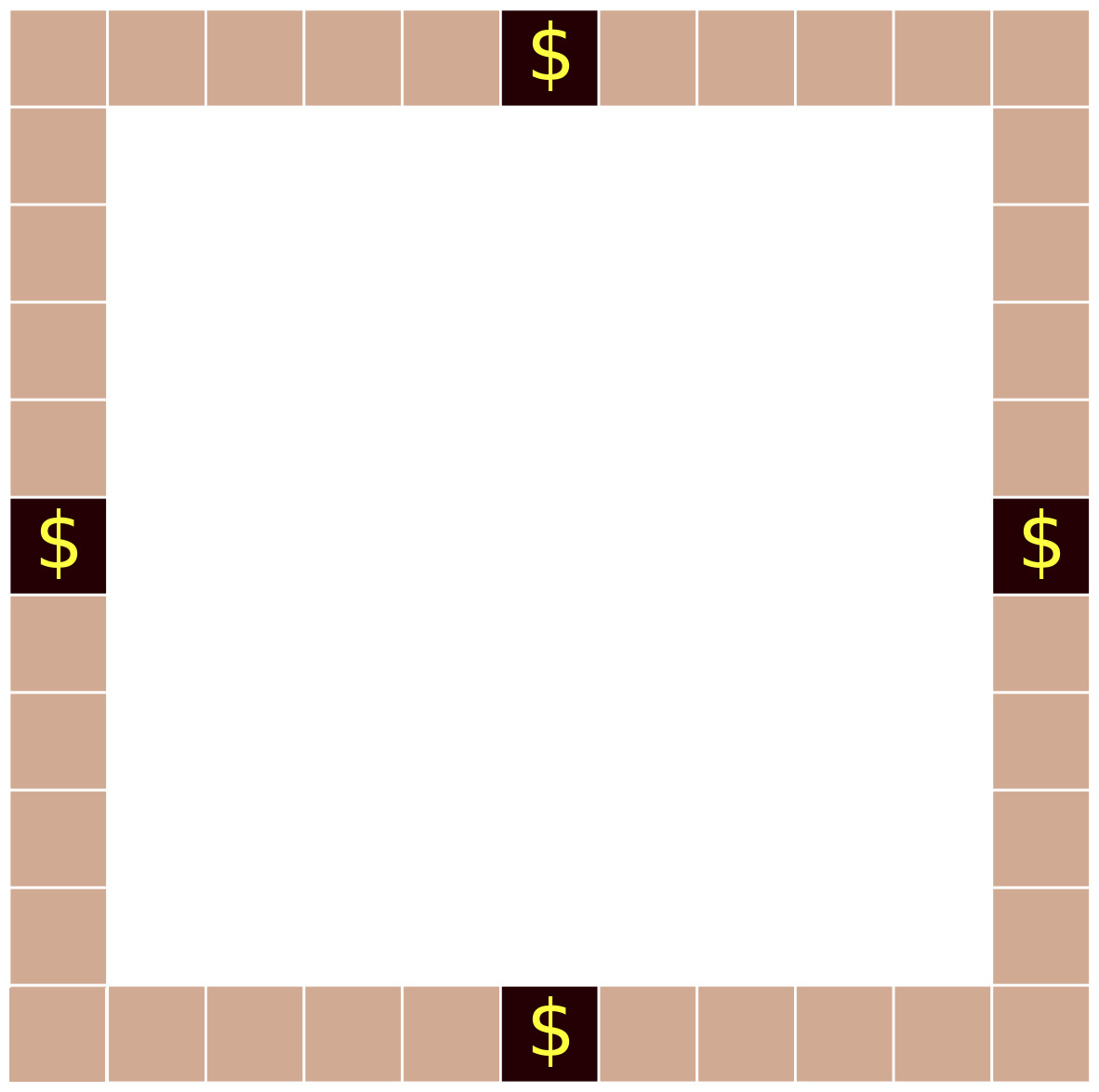}}
    \subfigure[HalfCheetah]{\label{fig:hc}\includegraphics[width=0.19\textwidth]{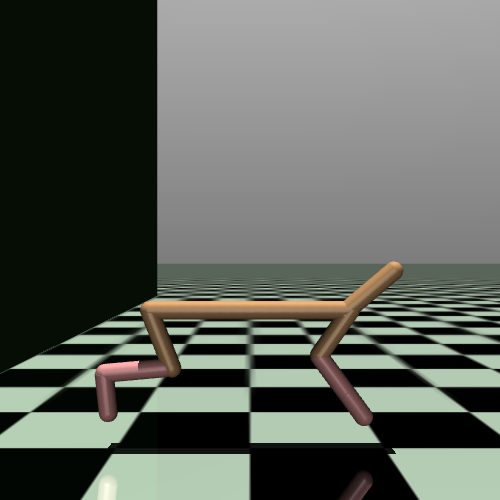}}
    \subfigure[Ant]{\label{fig:ant}\includegraphics[width=0.19\textwidth]{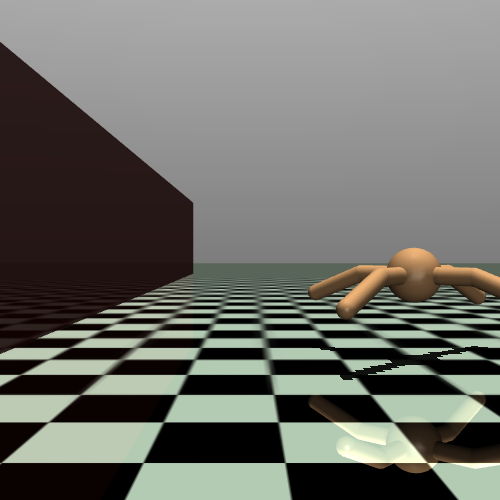}}
    \subfigure[Point]{\label{fig:point}\includegraphics[width=0.19\textwidth]{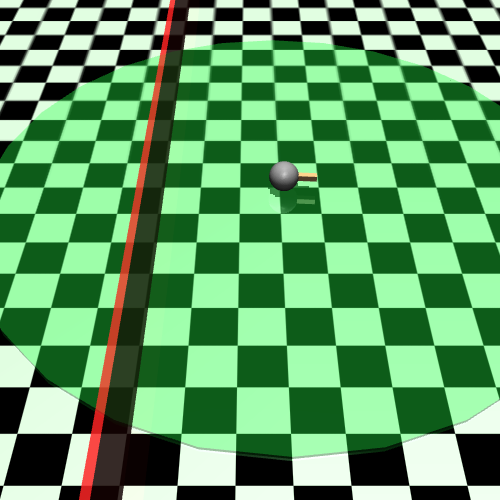}}
    \subfigure[AntBroken]{\label{fig:ant-broken}\includegraphics[width=0.19\textwidth]{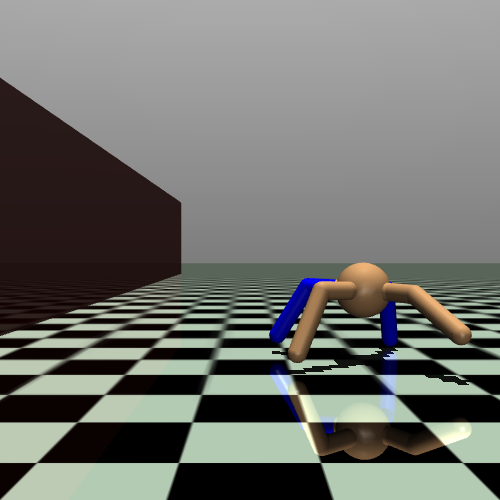}}
    \caption{The environments used in the experiments. Note that nominal agents are not aware of the constraints shown.}
    \label{fig:envs}
\end{center}
\vskip -0.2in
\end{figure*}

\subsection{KL-Based Early Stopping}
Importance sampling allows us to run multiple updates of $\zeta_\theta$ after each forward step. However, importance sampling weights may have high variance, which, in practice, may result in bad updates to $\zeta_\theta$ especially when policies $\pi_{\bar\theta}$ and $\pi_{\theta}$ become very ``different''. To mitigate this, we impose an upper bound on the forward and reverse KL divergences between $\pi_{\bar\theta}$ and $\pi_\theta$. We do this by terminating updates of $\zeta_\theta$ once the KL-divergences exceed some pre-defined threshold. The forward and reverse KL-divergences can be bounded in terms of the importance sampling weights as follows (see Section \ref*{a:kl_is} in the supplementary material for derivation).
\begin{equation}
\begin{split}
    D_\text{KL}(\pi_{\bar\theta}\vert\vert\pi_\theta) &\leq 2\log \bar\omega\\
    D_{KL}(\pi_{\theta}\vert\vert\pi_{\bar\theta}) &\leq \frac{\mathbb{E}_{\tau\sim\pi_{\bar\theta}} \left[(\omega(\tau)-\bar\omega)\log\omega(\tau)\right]}{\bar{\omega}}.
\end{split}
    \label{eq:trust_kl}
\end{equation}
Here $\bar\omega = \mathbb{E}_{\tau\sim\pi_{\bar\theta}}\left[\omega(\tau)\right]$ and $\omega(\tau) = \prod_{t=1}^T\omega(s_t,a_t)$.

Algorithm \ref{algo:1} summarizes our training procedure.

\begin{algorithm}[t]
    \caption{ICRL}
    \label{algo:1}
    \begin{algorithmic}
        \STATE {\bfseries Input:} Expert trajectories $\mathcal{D}$, iterations $N$, backward iterations $B$, maximum allowable KLs $\epsilon_F$ and $\epsilon_R$ 
        \STATE Initialize $\theta$ and $\phi$ randomly
        \FOR{$i=1$ {\bfseries to} $N$}
        \STATE Learn $\pi^\phi$ by solving (\ref{eq:min_max}) using current $\zeta_\theta$\\
        \FOR{$j=1$ {\bfseries to} $B$}
        \STATE Sample set of trajectories $\mathcal{S} = \{\tau^{(k)}\}_{k=1}^M$ from $\pi^\phi$\\
        \STATE Compute importance sampling weights $w(\tau^{(k)})$ using (\ref{eq:is_weights}) for $k=1,\ldots,M$\\
        \STATE Use $\mathcal{S}$ and $\mathcal{D}$ to update $\theta$ via SGD by using the gradient in (\ref{eq:grad_theta_is})
        \STATE Compute forward and reverse KLs using (\ref{eq:trust_kl})
        \IF{forward KL $\geq \epsilon_F$ {\bfseries or} reverse KL $\geq \epsilon_R$:}
        \STATE {\bfseries break}
        \ENDIF
        \ENDFOR
        \ENDFOR
    \end{algorithmic}
\end{algorithm}

\begin{figure*}[ht!]
\vskip 0.2in
\begin{center}
    \small{\bfseries Reward (higher is better):}\\
    \subfigure{\includegraphics[width=0.24\textwidth]{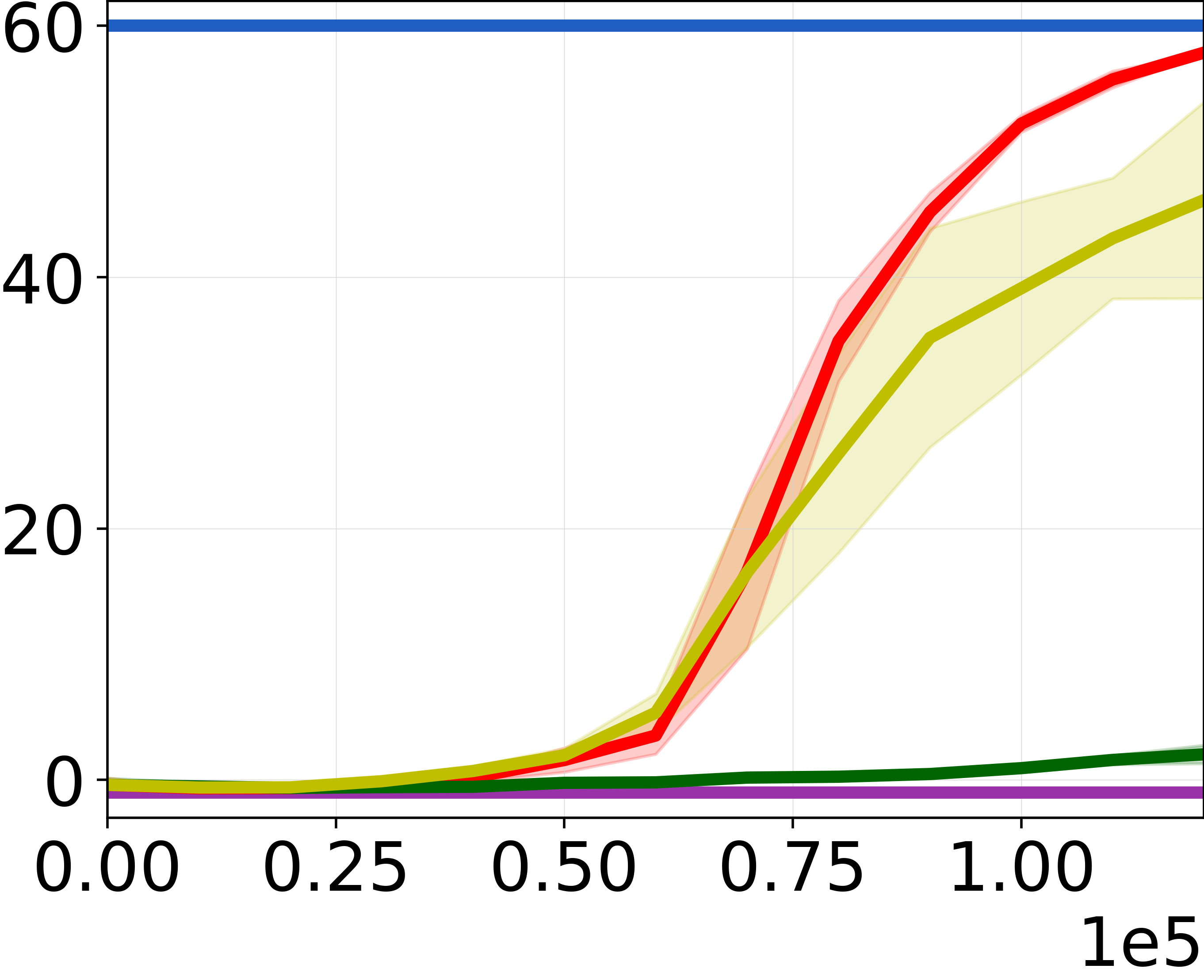}}\hspace{0.2in}
    \subfigure{\includegraphics[width=0.24\textwidth]{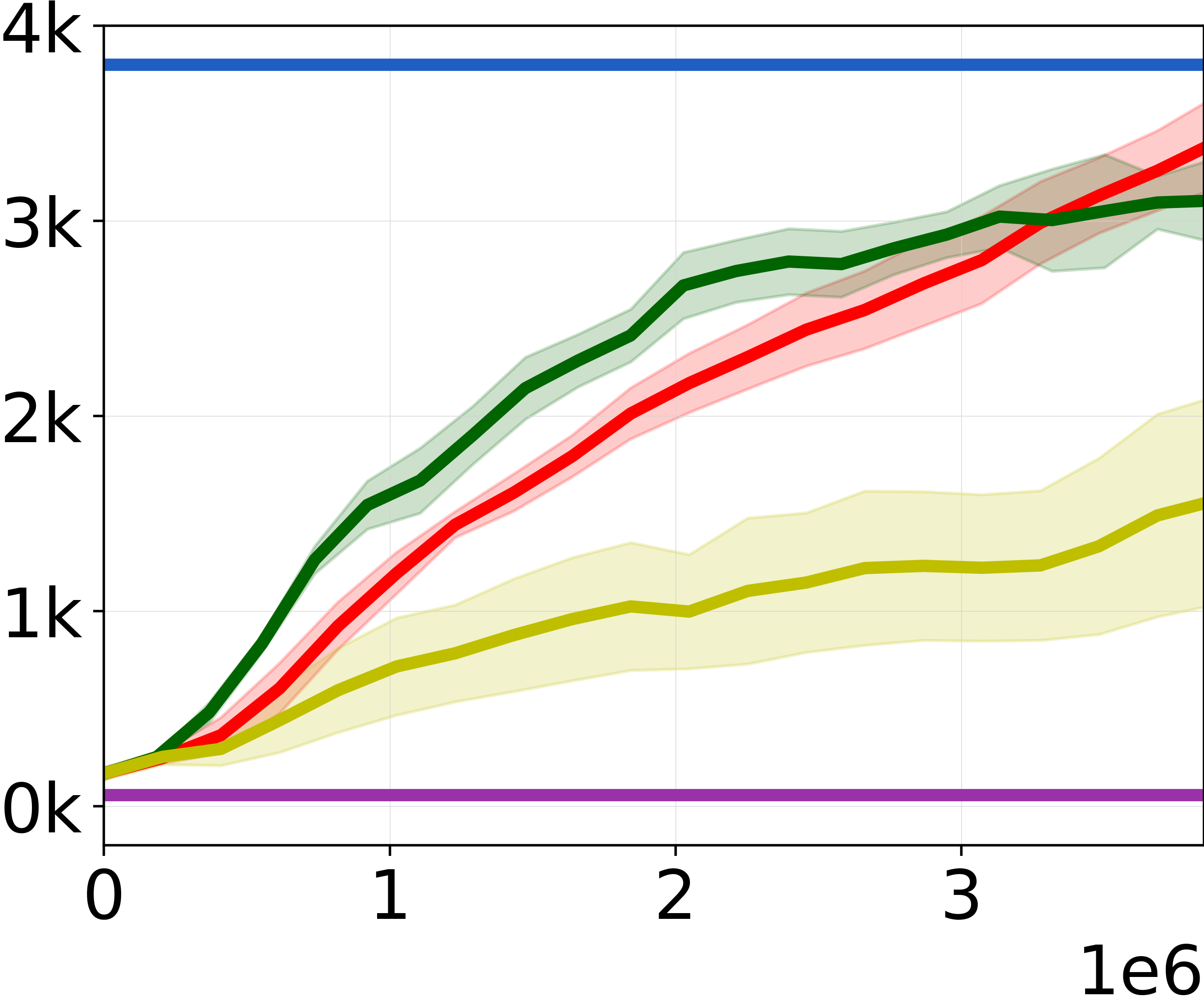}}\hspace{0.2in}
    \subfigure{\includegraphics[width=0.24\textwidth]{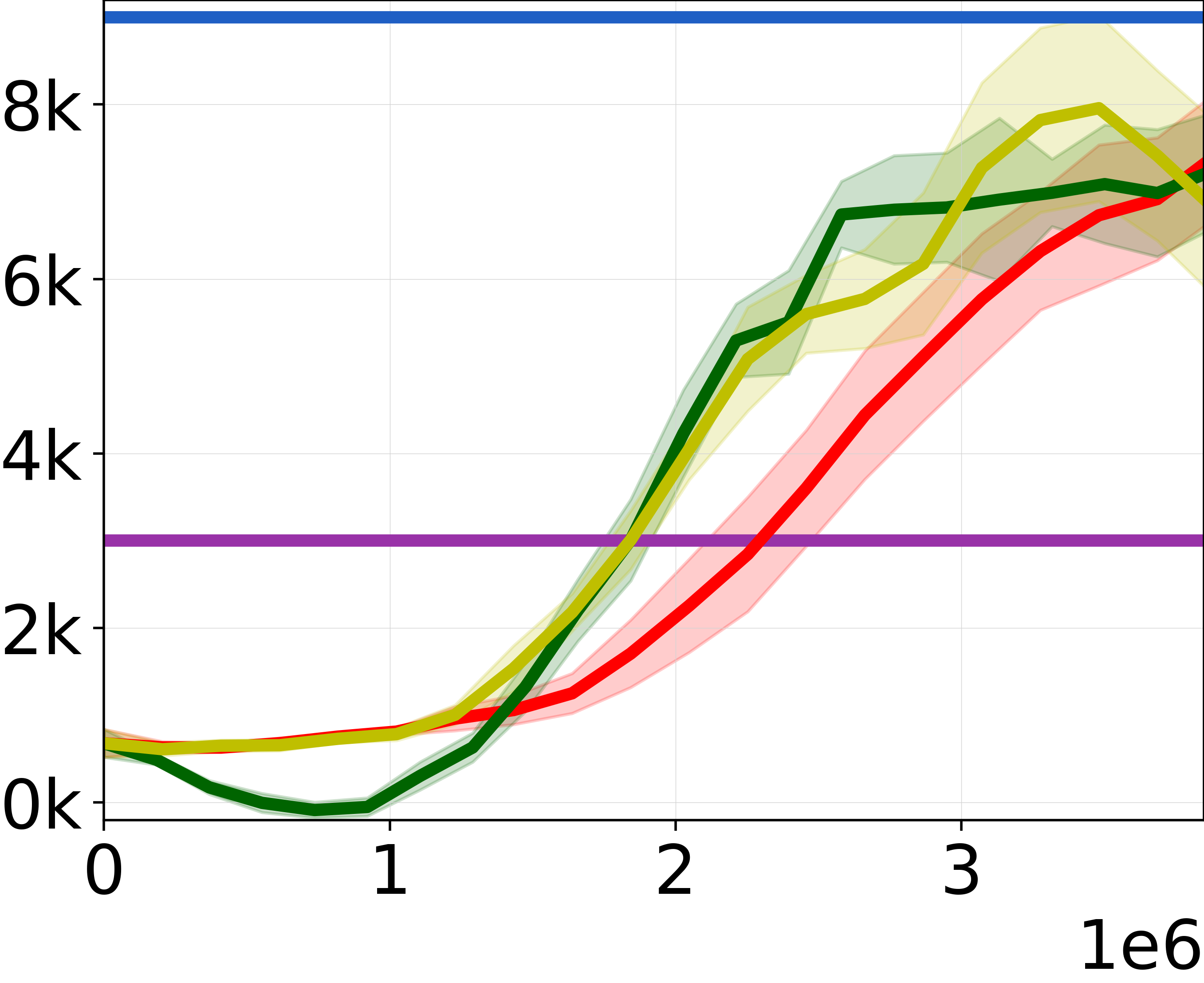}}\\
    \setcounter{subfigure}{0}
    \small{\bfseries Constraint violations (lower is better):}\\
    \subfigure[LapGridWorld]{\includegraphics[width=0.24\textwidth]{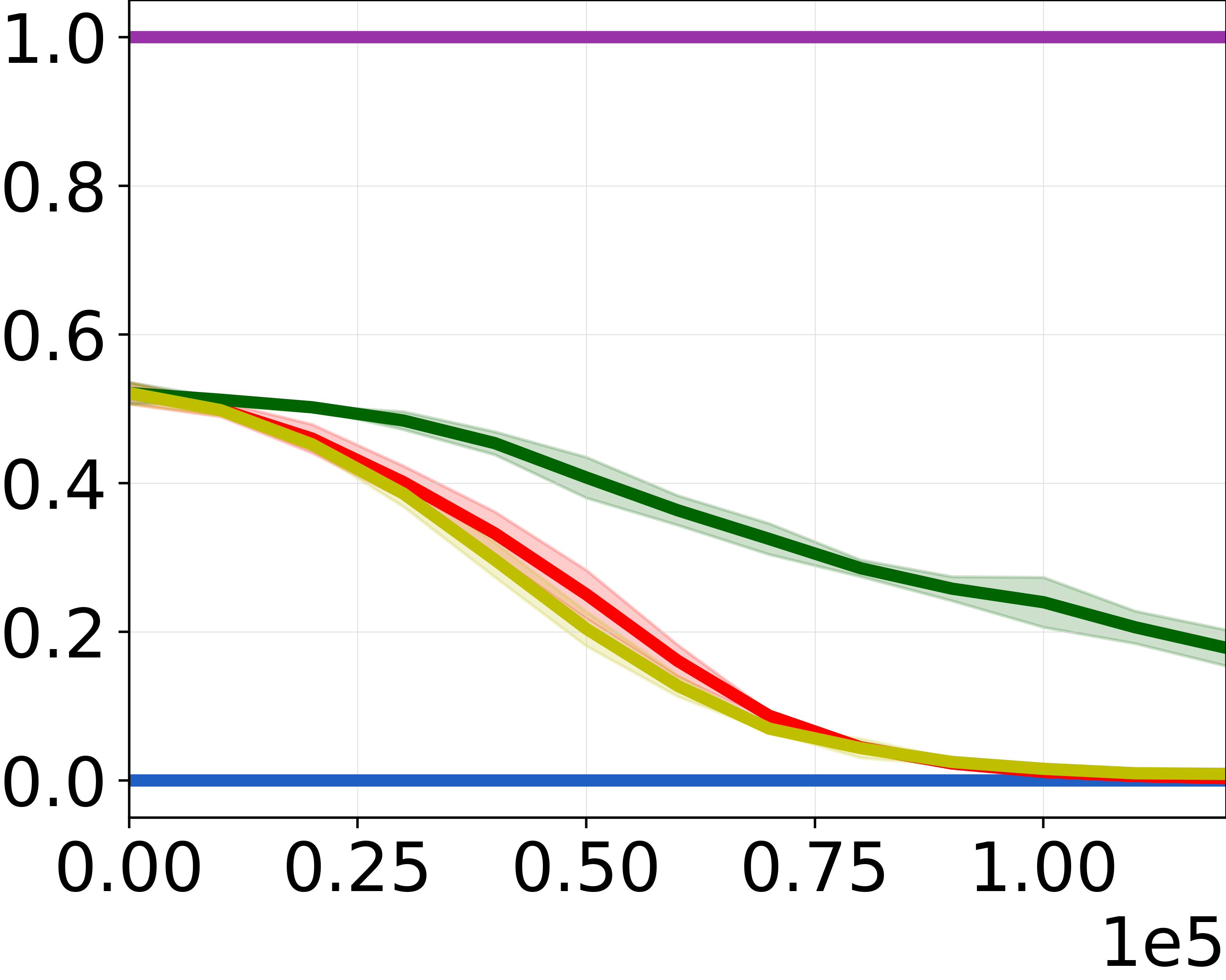}}\hspace{0.2in}
    \subfigure[HalfCheetah]{\includegraphics[width=0.24\textwidth]{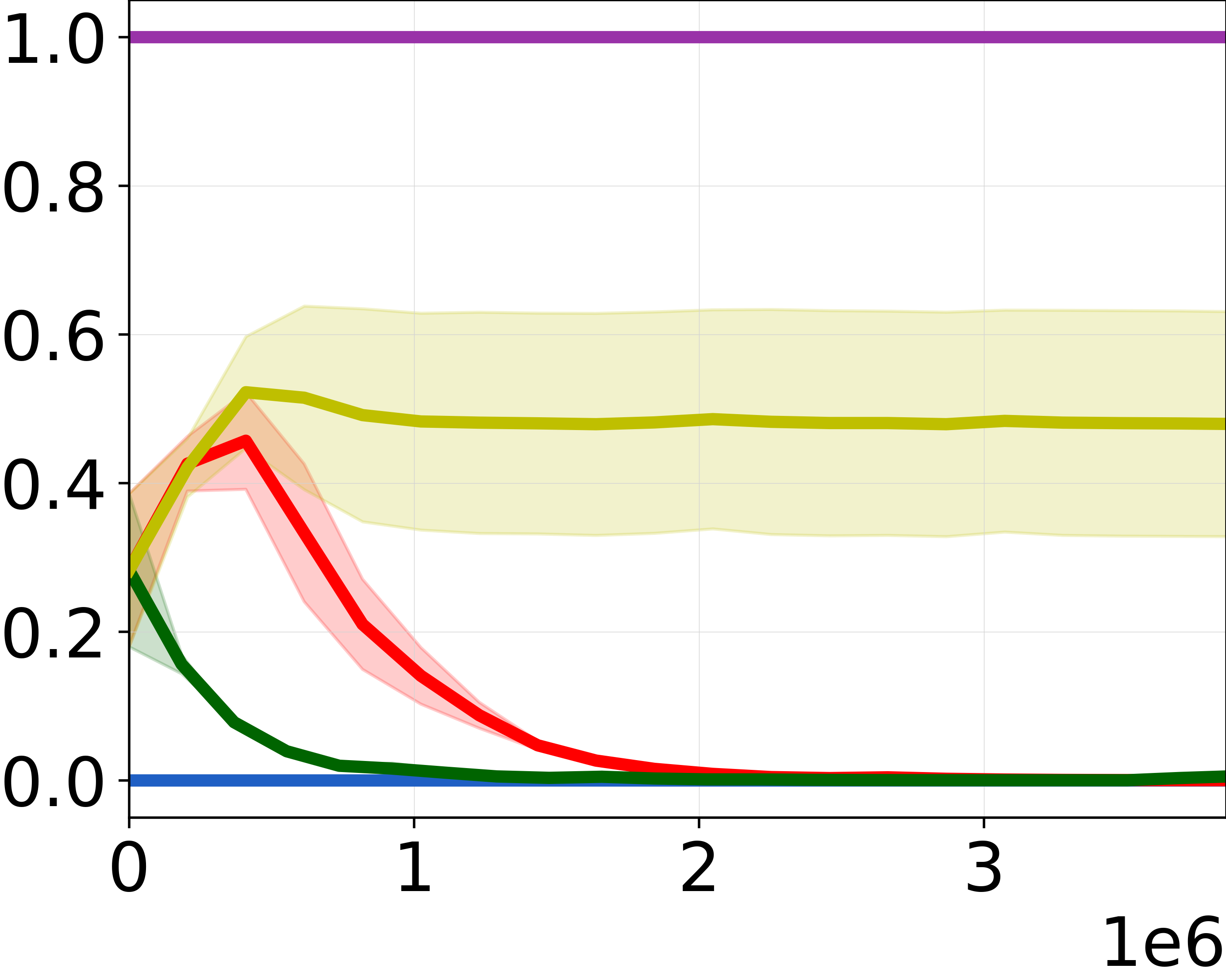}}\hspace{0.2in}
    \subfigure[Ant]{\includegraphics[width=0.24\textwidth]{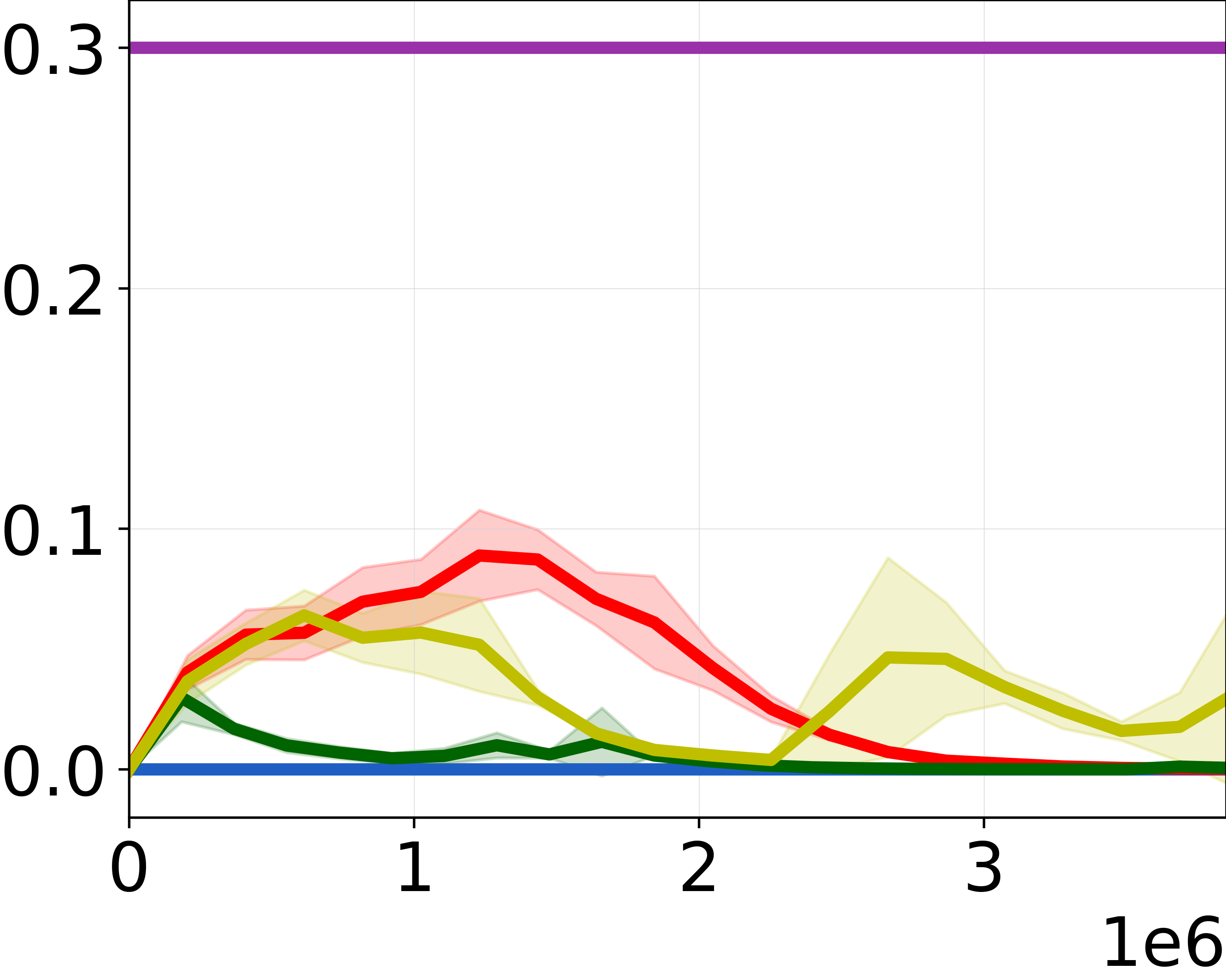}}
    \subfigure{\includegraphics[width=0.6\textwidth]{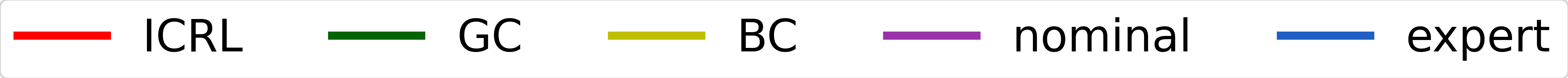}}\\
    \caption{Performance of agents during training over several seeds (5 in LapGridWorld, 10 in others). The x-axis is the number of timesteps taken in the environment. The shaded regions correspond to the standard error.}
    \label{fig:main_results}
\end{center}
\vskip -0.2in
\end{figure*}

\begin{figure}[t]
\vskip 0.2in
\begin{center}
    \subfigure{\includegraphics[width=0.49\columnwidth]{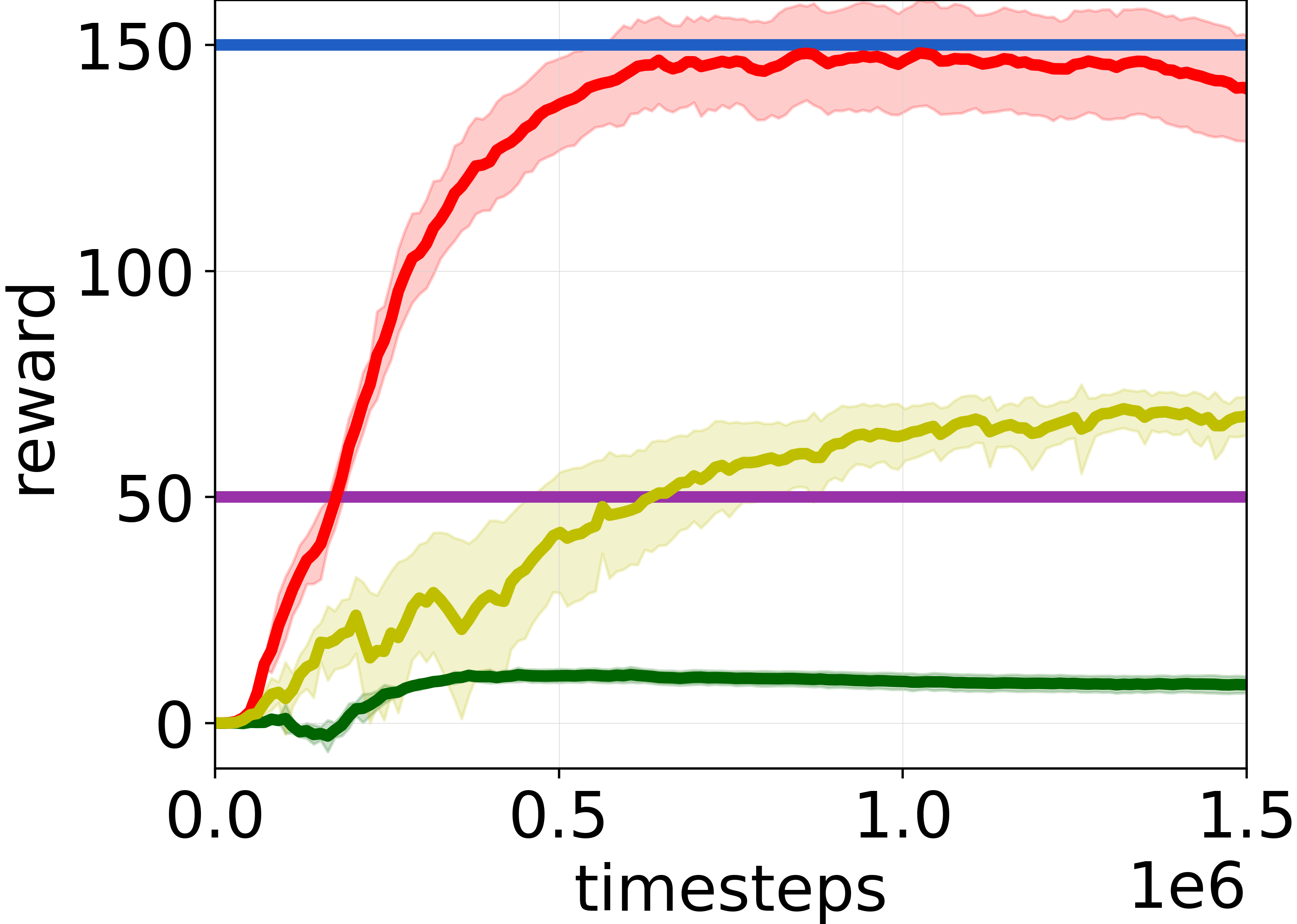}}
    \subfigure{\includegraphics[width=0.49\columnwidth]{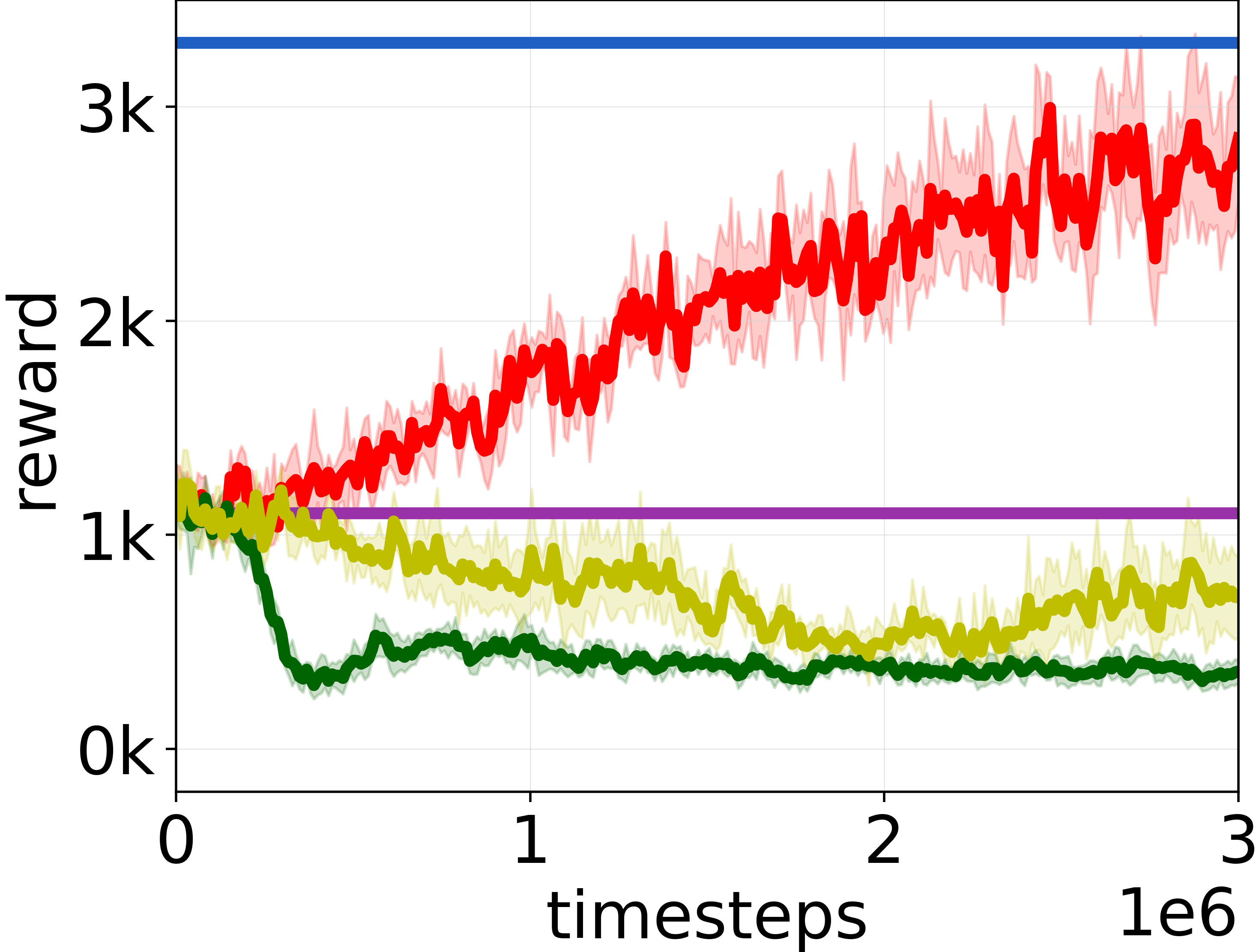}}\\
    \setcounter{subfigure}{0}
    \subfigure[Point]{\includegraphics[width=0.49\columnwidth]{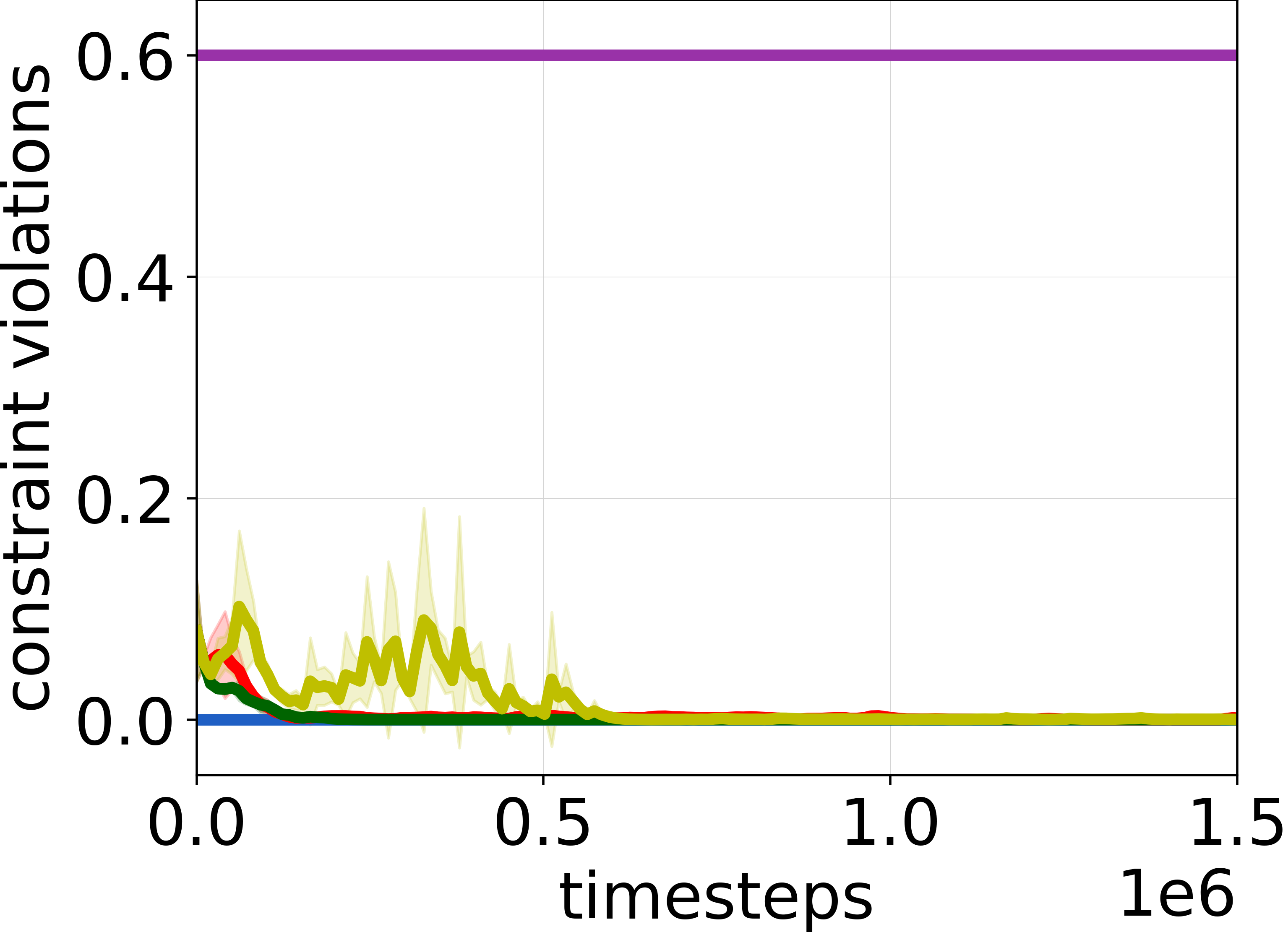}}
    \subfigure[Ant-Broken]{\includegraphics[width=0.49\columnwidth]{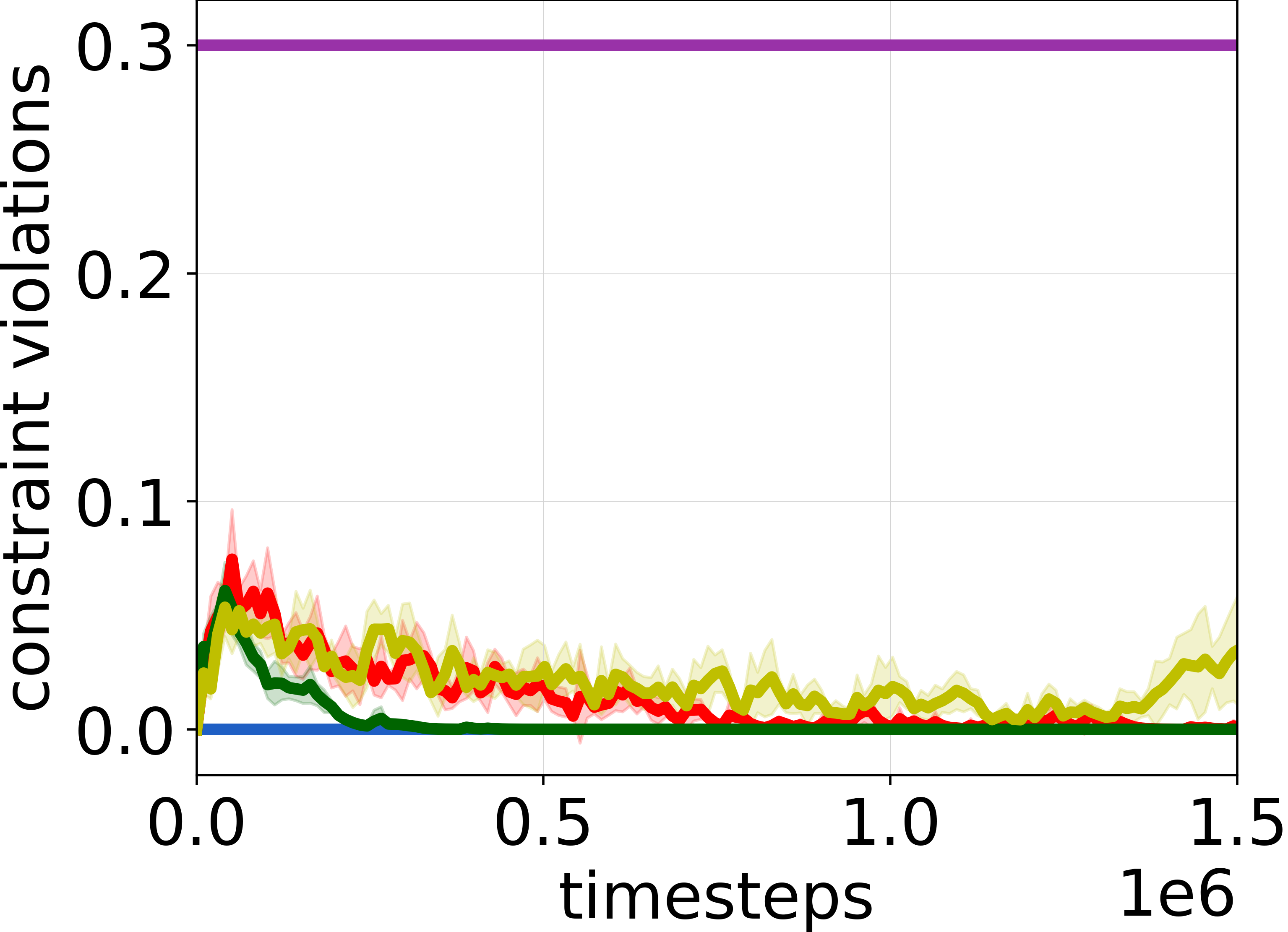}}
    \subfigure{\includegraphics[width=\columnwidth]{figures/legend.png}}\\
    \caption{Transferring constraints. The x-axis is the number of timesteps taken in the environment. All plots were smoothed and averaged over 5 seeds. The shaded regions correspond to the standard error.}
    \label{fig:transfer}
\end{center}
\vskip -0.2in
\end{figure}

\section{Experiments}
\label{sec:exp}
We conduct experiments to validate our method and assess the quality of constraints it recovers. In the following discussion, we will refer to both the nominal and constrained environments. The nominal environment corresponds to the simulator we use to train our agent (also called as the nominal agent). It does not include any information about the constraints. The constrained (or true) environment refers to the real-world setting where episodes are terminated if the agent violates any constraints. We report two metrics for each experiment that we conduct: (1) the true reward which corresponds to the reward accumulated by the agent in the constrained environment, and (2) the average number of constraint violations per each timestep. For the latter metric, we assume that the true constraints are known. However, the training algorithm does not require this information. Note that since episodes get terminated in the constrained environment as soon as an agent violates some constraint, agents simply trained to maximize the reward in the nominal environment will only be able to accumulate a small amount of reward in the constrained environment.

Furthermore, we construct the following two baselines for comparisons:
\begin{itemize}
    \item {\bfseries Binary Classifier (BC):} We simply train a binary classifier (using the cross-entropy loss) to classify between nominal and expert trajectories. We train the agent by solving, as before, (\ref{eq:min_max}) (note that the binary classifier corresponds to $\zeta$).
    \item {\bfseries GAIL-Constraint (GC):} This baseline is based on a well-known imitation learning method called GAIL \citep{ho2016gail}. Since GAIL does not assume knowledge of the reward function, while our method does, we construct the following variant for a fairer comparison. We begin by defining the following reward function: $\bar{r}(s,a) := r(s,a) + \log \zeta(s,a)$. Note that since for constrained state-action pairs $\bar{r}$ will be $-\infty$, agents maximizing $\bar{r}$ will try to satisfy all constraints. Now since we already know $r$, we only parameterize $\zeta$ as a neural network which, then, corresponds to the GAIL's discriminator.
\end{itemize}

We design our experiments to answer the following two questions:
\begin{itemize}
    \item Can ICRL train agents to abide by the constraints that an expert does?
    \item Are the constraints recovered by ICRL transferable to other agents with possibly different morphologies and/or reward functions?
\end{itemize}
As we show below, the answer to both of these questions is in the affirmative. Furthermore, while the GC baseline performs equally well to ICRL in the first case, it completely fails in the second one.

Finally, all hyperparameters and additional details of the experiments can be found in Section \ref*{a:es} in the supplementary material.

\subsection{Learning Constraints}
In this section, we design experiments to test whether ICRL can train agents to abide by the constraints that an expert agent does.

\textbf{Reward hacking:} Reward hacking refers to situations in which agents optimize their rewards but their behavior is misaligned with the intentions of the users, e.g., when a vaccum cleaner ejects dust so that it can collect even more \citep{russel2010ai}.  We design a simple environment called LapGridWorld to see if our method can avoid this behavior.. In this environment\footnote{This environment is similar to the boat race environment in \citet{leike2017gridworlds}.} (shown in Figure \ref{fig:lapgw}) an agent moves on a rectangular track. The agent is \textit{intended} to sail clockwise around the track. Each time it drives onto a golden dollar tile, it gets a reward of 3. However, the nominal agent ``cheats'' by stepping back and forth on one dollar tile, rather than going around the track, and ends up getting more reward than the expert (which goes around the track, as intended).

\textbf{High dimensions:} For this experiment, we use two simulated robots called HalfCheetah and Ant from OpenAI Gym \citep{brockman2016openai}. The robots are shown in Figures \ref{fig:hc} and \ref{fig:ant} respectively. The HalfCheetah robot has $18$ state and $6$ action dimensions and is rewarded proportionally to the distance it covers in each step. The Ant robot has $113$ state and $8$ action dimensions and is rewarded proportionally to the distance it moves away from the origin. Since, for both agents, moving backwards is relatively easier than moving forwards, in the constrained environments, we constrain all states for which $x\leq -3$.

We plot the results of our experiments in Figures \ref{fig:main_results}. The x-axis corresponds to the number of timesteps the agent takes in the environment. The top row corresponds to the reward of the agent during training when evaluated in the constrained environment. The bottom row shows the average number of constraints that the agent violates per timestep (when run in the nominal environment). In addition to our method (ICRL) and the two baselines, we also show the reward and average number of constraint violations of the nominal and expert agents. As can be observed, both ICRL and GC perform equally well and are able to achieve expert-level performance. However, BC performs performs much more worse than these two.

\subsection{Transferring constraints} 
In many cases, constraints are actually part of the environment and are the same for different agents (for example, \textbf{all} vehicles must adhere to the same speed limit). In such instances, it is useful to first learn the constraints using only one agent and then transfer them onto other agents. These new agents may have different morphologies and/or reward functions. In this section, we design experiments to assess the quality of constraints recovered by our method and the two baselines. To do so, we try to transfer constraints learnt in the Ant experiment in the previous section to two new agents. The first agent, shown in Figure \ref{fig:point}, is the Point robot from OpenAI gym. The Point robot's reward function encourages it to move counter-clockwise in a circle at a distance of $10$ units from the origin.\footnote{This environment is similar to the PointCircle environment in \citet{achiam2017cpo}.} The second agent, called Ant-Broken, is the Ant robot with two of its legs disabled (colored in blue in Figure \ref{fig:ant-broken}).\footnote{This is based on the broken ant environment in \citet{eysenbach2020off}.} To transfer constraints, we simply solve (\ref{eq:min_max}) using the $\zeta$ learnt in the previous section.\footnote{Note that in this case, $\zeta $ must only be fed a subset of the state and action spaces that are common across the two agents. As such, for the Point experiment we re-run experiments in the previous section and train $\zeta$ only on the $(x,y)$ position coordinates of the Ant agent, since the rest of elements in the state space are specific to Ant.}

Figure \ref{fig:transfer} shows the results. The plots represent the same metrics as in the previous section. Note that only agents trained to respect constraints recovered by our method are able to achieve expert-level performance. Both BC and GC perform even worse in terms of the reward than the nominal agent, which is simply trained to maximize the reward in the nominal environment without worrying about any constraints. However, note that both BC and GC are able to achieve zero cost. This suggests that both BC and GC constrain too many state-action pairs, thus preventing agents from performing their tasks optimally.

\begin{figure*}[h!]
\vskip 0.2in
\begin{center}
    \small{\bfseries Reward (higher is better):}\\
    \subfigure{\includegraphics[width=0.24\textwidth]{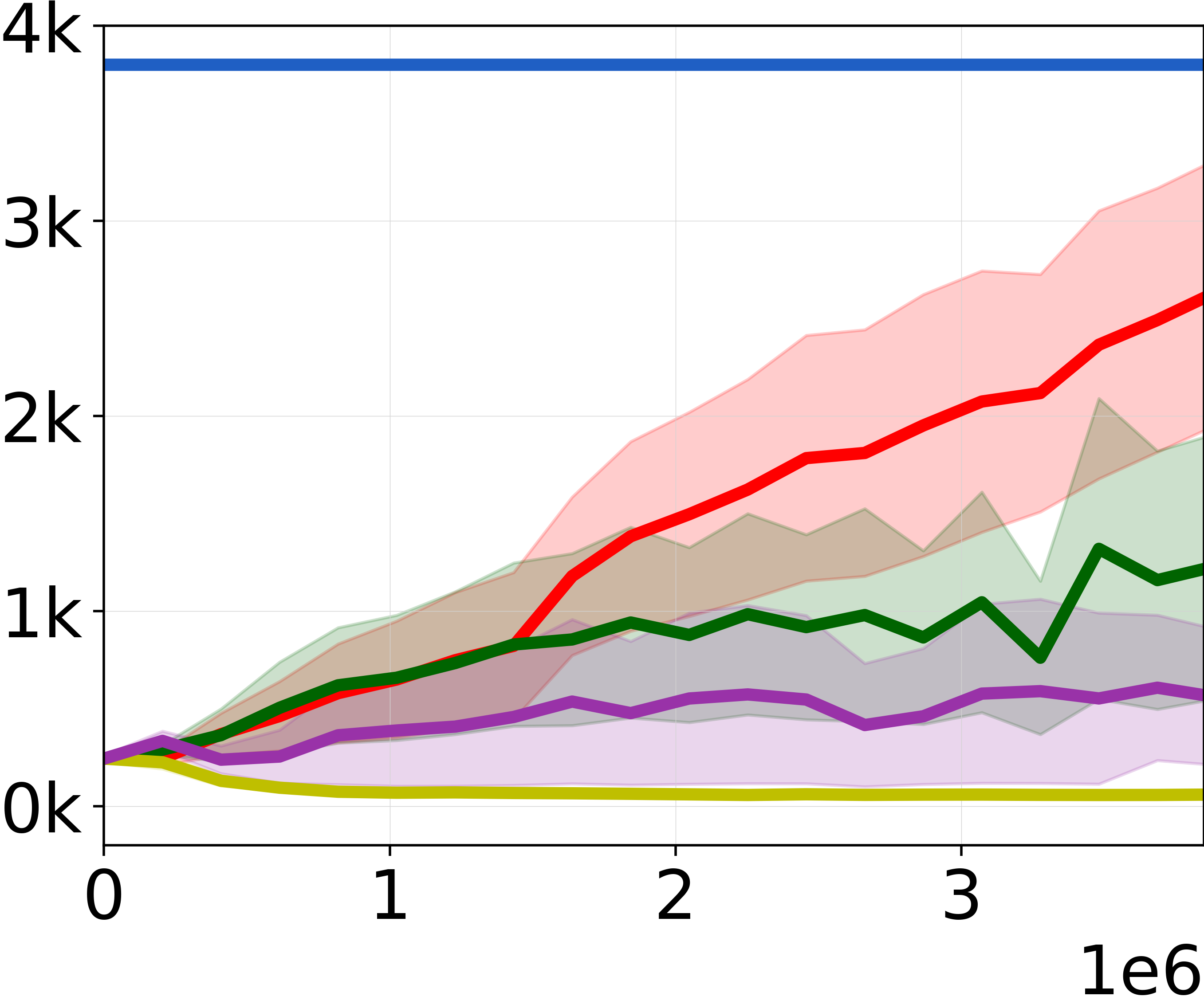}}
    \subfigure{\includegraphics[width=0.24\textwidth]{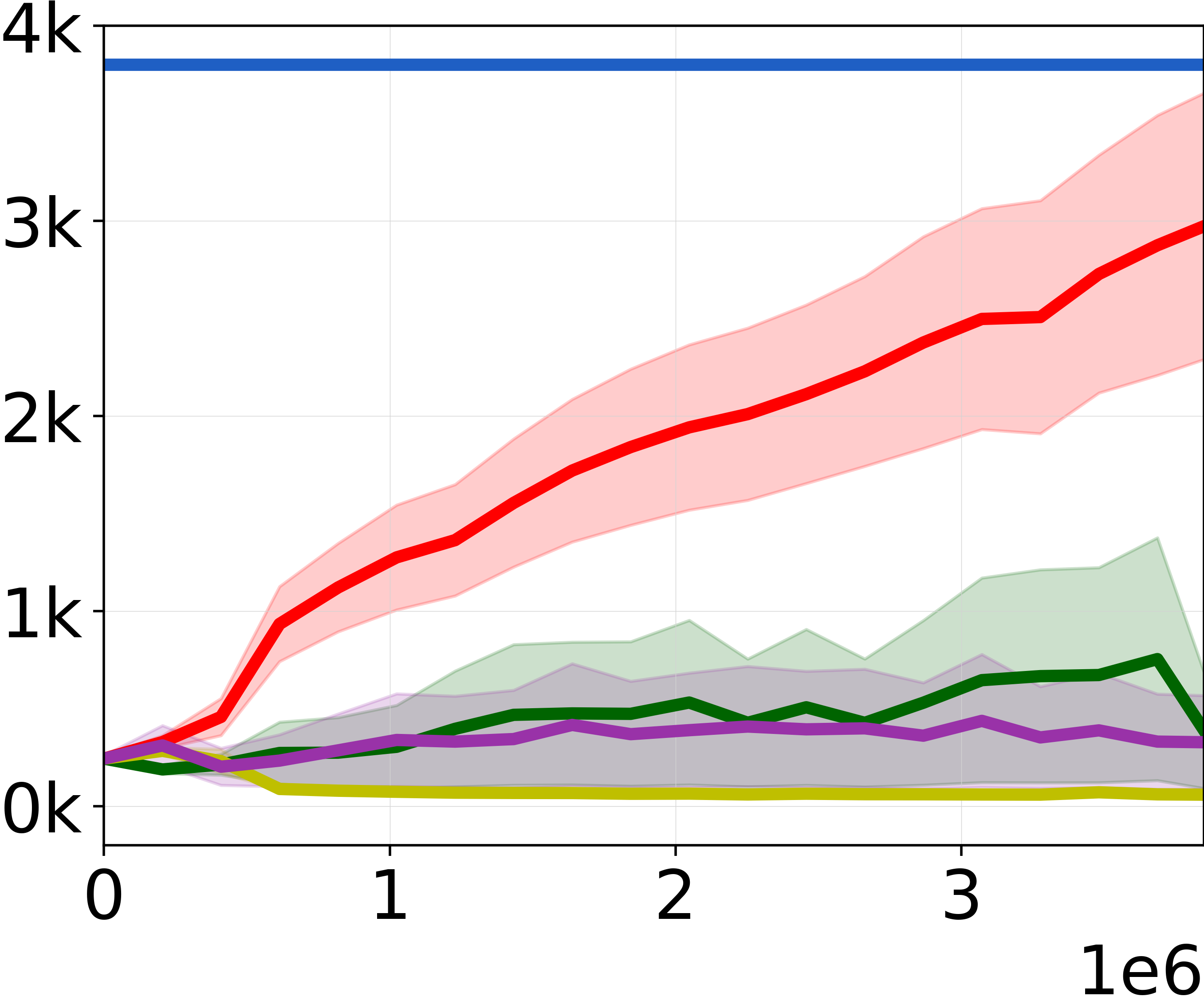}}
    \subfigure{\includegraphics[width=0.24\textwidth]{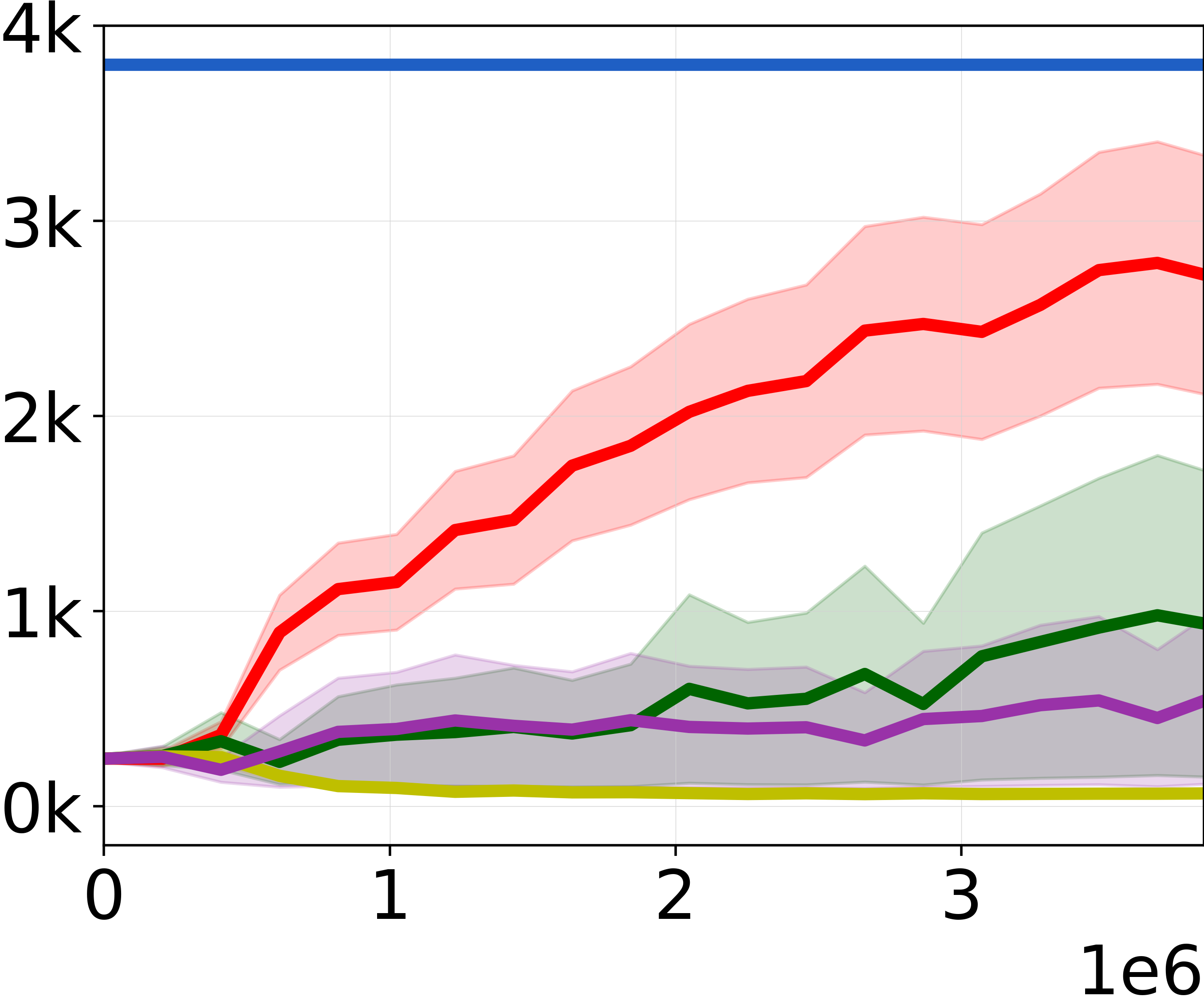}}
    \subfigure{\includegraphics[width=0.24\textwidth]{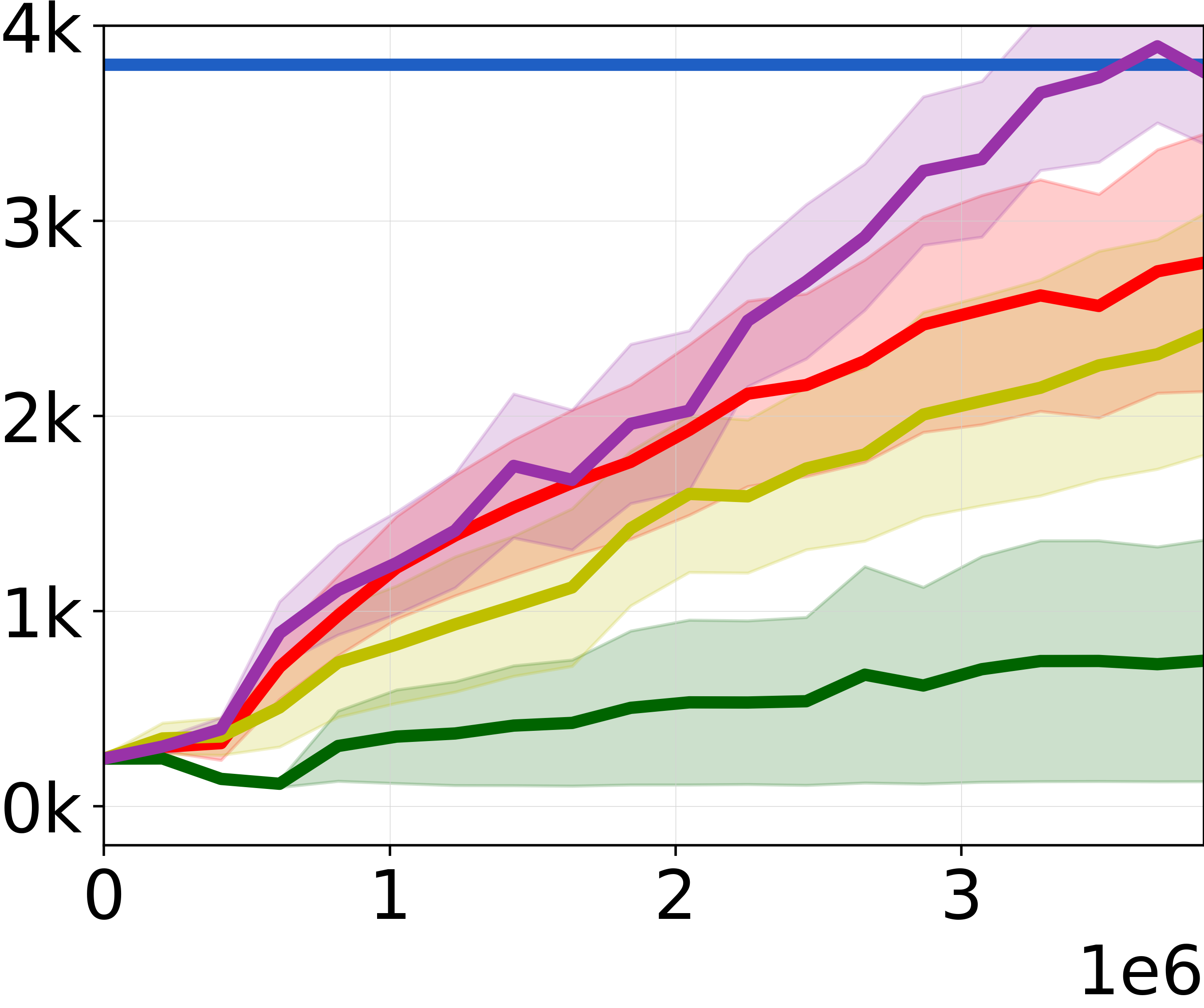}}\\
    \setcounter{subfigure}{0}
    \small{\bfseries Constraint violations (lower is better):}\\
    \subfigure[No IS and no ES]{\includegraphics[width=0.24\textwidth]{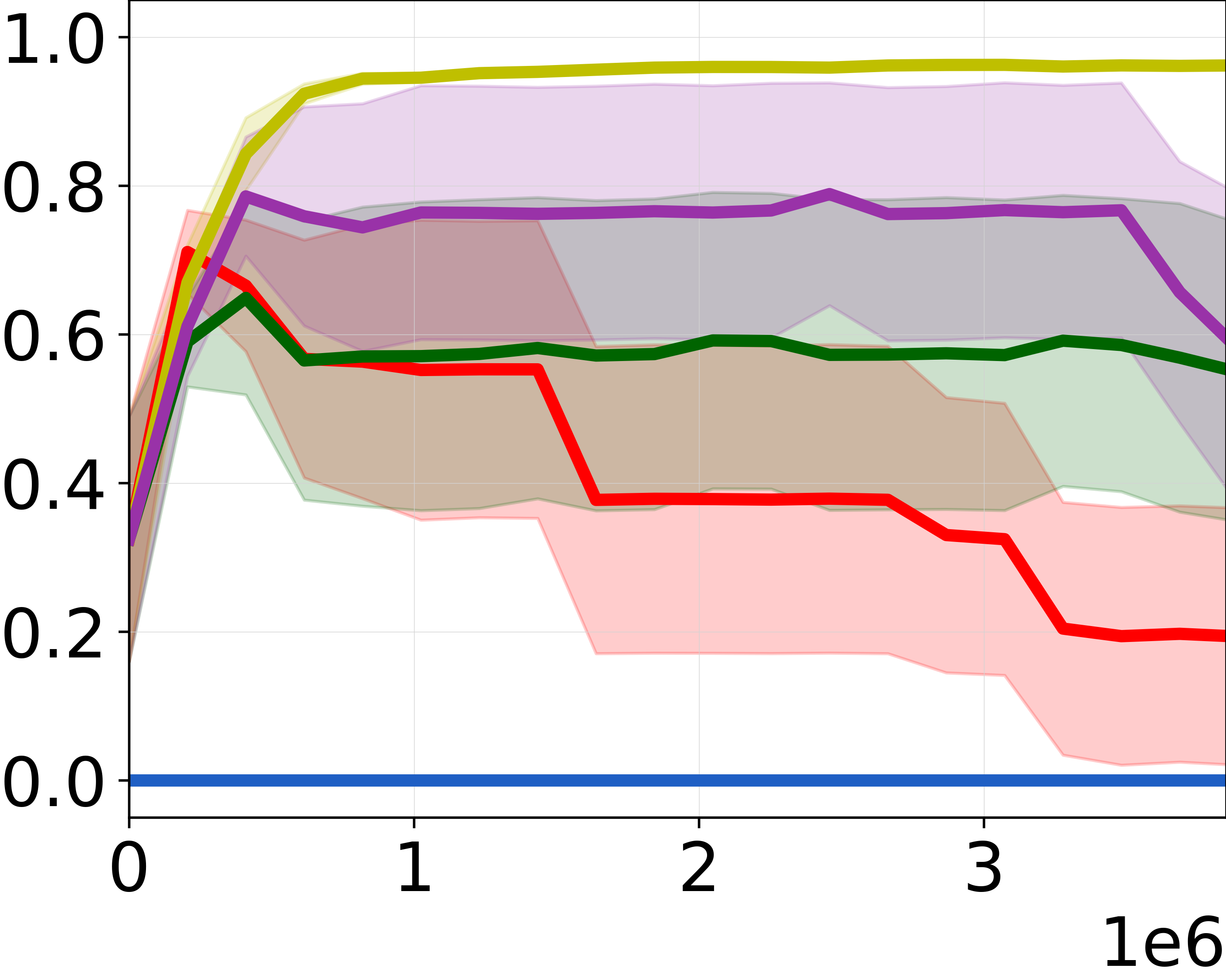}}
    \subfigure[No IS but ES]{\includegraphics[width=0.24\textwidth]{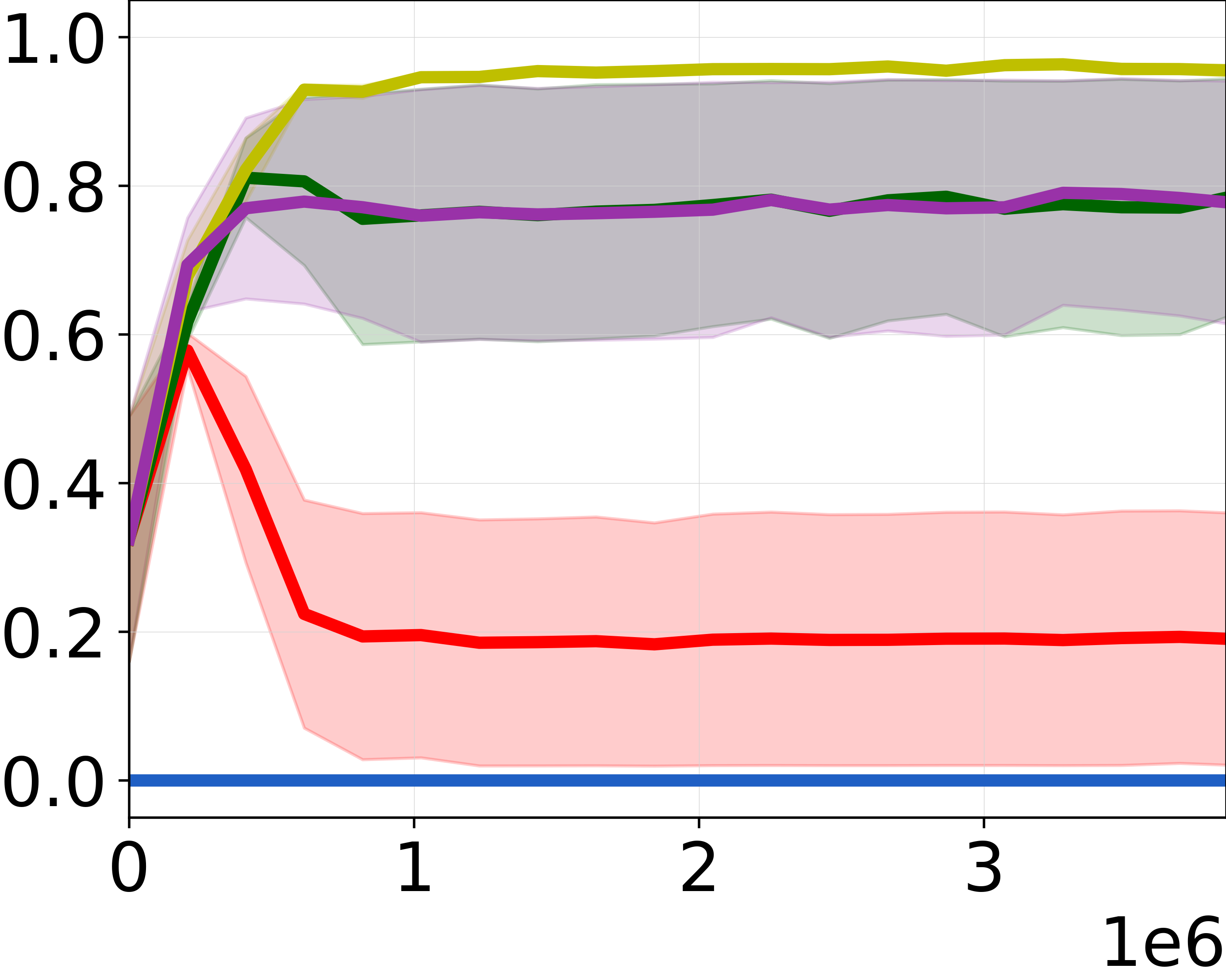}}
    \subfigure[IS but no ES]{\includegraphics[width=0.24\textwidth]{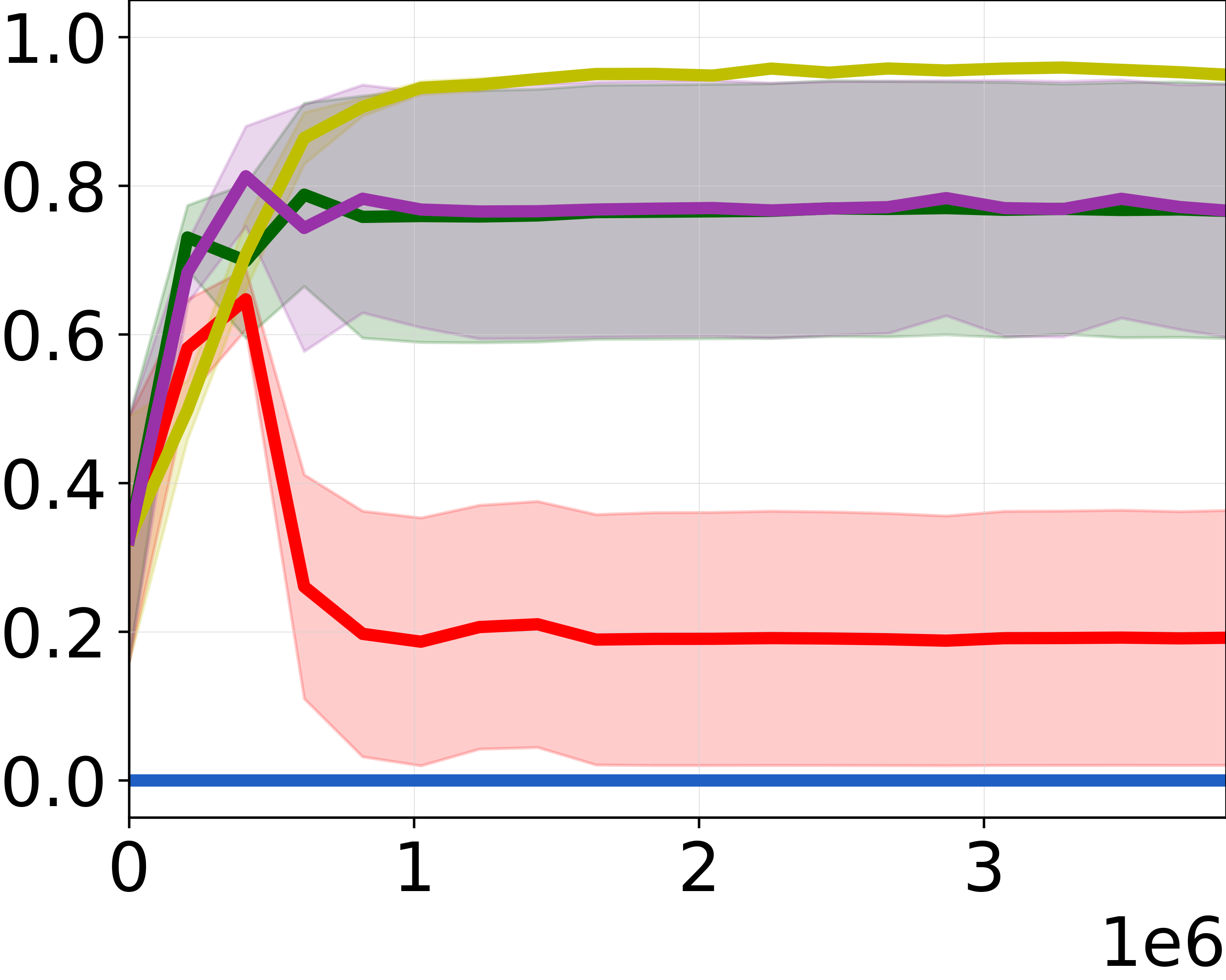}}
    \subfigure[IS and ES]{\includegraphics[width=0.24\textwidth]{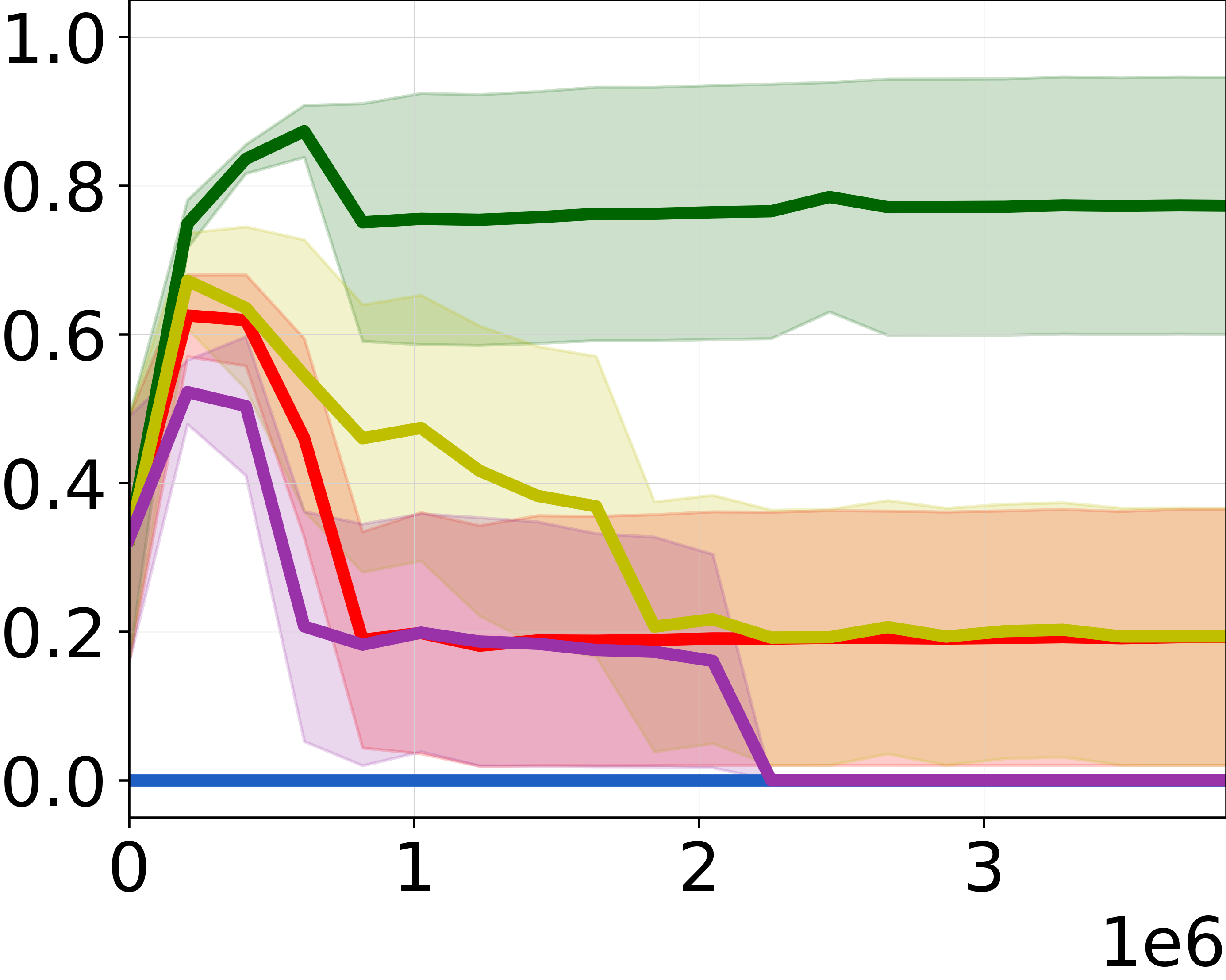}}\\
    \subfigure{\includegraphics[width=0.6\textwidth]{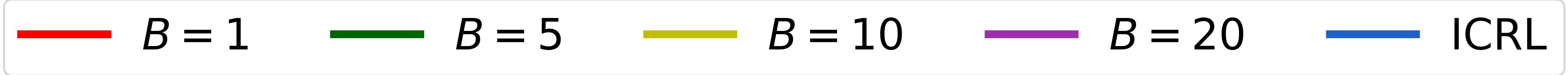}}\\
    \caption{Ablation studies on the HalfCheetah environment. All plots were averaged over $5$ seeds. IS refers to importance sampling and ES to early stopping. The x-axis corresponds to the number of timesteps the agent takes in the environment. Shaded regions correspond to the standard error.}
    \label{fig:abl}
\end{center}
\vskip -0.2in
\end{figure*}

\section{Ablation Studies}
In this section, we discuss ablation studies to assess the importance of importance sampling and the KL-based early stopping technique. We carry out four sets of experiments by disabling either one or both of importance sampling and early stopping. Additionally, we also plot results for the case when neither of importance sampling and early stopping are disabled. In each set of experiments, we vary the number of backward iterations, $B$ (see Algorithm \ref{algo:1} and observe (as in the previous section) the true reward (on the constrained environment) and average number of constraint violations. All four sets of experiments are shown in Figure \ref{fig:abl}. For comparison, we also show the best true reward and lowest number of constraint violations that our method was able to achieve with all of its components enabled (these are the same results as in the previous section).

From the plots, it is clear that both of importance sampling and early stopping are important to the success of our method.

\section{Related Work}
\textbf{Forward Constrained RL:} Several approaches have been proposed in literature to solve the forward constrained RL problem in the context of CMDPs \citep{altman1999cmdps}. \citet{achiam2017cpo} analytically solves trust region policy optimization problems at each policy update to enforce the constraints. \citet{chow2018lyapunov} uses a Lyapunov approach and also provide theoretical guarantees. \citet{le2019batch} proposes an algorithm for cases when there are multiple constraints. Finally, a well-known approach centers around rewriting the constrained RL problem as an equivalent unconstrained min-max problem by using Lagrange multipliers \citep{zhang2019recommender, tessler2018reward, bhatnagar2010ac-lagrangian}) (see Section \ref{sec:fs} for further details).

\textbf{Constraint Inference:} Previous work done on inferring constraints from expert demonstrations has either focused on either inferring specific type of constraints such as geometric \citep{arpino2017geometric, subramani2018inferring}, sequential \citep{pardowitz2005sequential} or convex \citep{miryoosefi2019convexconstraints} constraints or is restricted to tabular settings \citep{scobee2020maximum, chou2018learning} or assumes transition dynamics \citep{chou2020local}.

\textbf{Preference Learning:} Constraint inference also links to preference learning which aims to extract user preferences (constraints imposed by an expert on itself, in our case) from different sources such as ratings \citep{danielP2014active}, comparisions \citep{chris2017preferences, sadigh2017active}, human reinforcement signals \citep{mac2017interactive} or the initial state of the agent's environment \citep{shah2019implicit}. Preference learning also includes inverse RL, which aims to recover an agent's reward function by using its trajectories. To deal with the inherent ill-posedness of this problem, inverse RL algorithms often incorporate regularizers \citep{ho2016gail, finn2016gcl} or assume a prior distribution over the reward function \citep{wonseok2018bayesiangail, michini2012nonparametric, ramachandran2007bayesianirl}.

\section{Conclusion and Future Work}
We have presented a method to learn constraints from an expert's demonstrations. Unlike previous works, our method both learns arbitrary constraints and can be used in continuous settings.

While we consider our method to be an important first step towards learning arbitrary constraints in real-world continuous settings, there is still considerable room for improvement. For example, as is the case with \citeauthor{scobee2020maximum}, our formulation is also based on (\ref{eq:maxent}) which only holds for deterministic MDPs. Secondly, we only consider hard constraints. Lastly, one very interesting extension of this method can be to learn constraints \textit{only} from logs data in an offline way to facilitate safe-RL in settings where it is difficult to even build nominal simulators such as is the case for plant controllers.

\bibliography{example_paper}
\bibliographystyle{icml2021}

\newpage
\onecolumn
\section{Appendix}
This appendix should be read as a continuation of the main paper.
\subsection{Gradient of log likelihood}
\label{a:gll}
The gradient of (\ref{eq:ll}) is
\begin{equation}
    \begin{split}
        \nabla_\theta\mathcal{L}(\theta) &= \frac{1}{N}\sum_{i=1}^M\left[0 + \nabla_\theta \log \zeta_\theta(\tau^{(i)})\right] - \frac{1}{\int\exp(\beta r(\tau))\zeta_\theta(\tau)d\tau}\int\exp(\beta r(\tau))\nabla_\theta \zeta_\theta(\tau)d\tau\\
        &= \frac{1}{N}\sum_{i=1}^N\nabla_\theta \log \zeta_\theta(\tau^{(i)}) - \int\frac{\exp(\beta r(\tau))\zeta_\theta(\tau)}{\int\exp(\beta r(\tau'))\zeta_\theta(\tau')d\tau'}\nabla_\theta \log \zeta_\theta(\tau)d\tau\\
        &= \frac{1}{N}\sum_{i=1}^N\nabla_\theta \log \zeta_\theta(\tau^{(i)}) - \int p_{\mathcal{M}^{\bar{\zeta}_\theta}}(\tau)\nabla_\theta \log \zeta_\theta(\tau)d\tau\\
        &= \frac{1}{N}\sum_{i=1}^N\nabla_\theta \log \zeta_\theta(\tau^{(i)}) - \mathbb{E}_{\tau \sim \pi_{\mathcal{M}^{\zeta_\theta}}}\left[\nabla_\theta \log \zeta_\theta(\tau)\right],
    \end{split}
\end{equation}
where the second line follows from the identity $\nabla_\theta \zeta_\theta(\tau) \equiv \zeta_\theta(\tau) \nabla_\theta \log \zeta_\theta(\tau)$ and the fourth line from the MaxEnt assumption.

\subsection{Derivation of importance sampling weights}
\label{a:is}
Suppose that at some iteration of our training procedure we are interested in approximating the gradient of the log of the partition function $\nabla_\theta \log Z_\theta$ (where $\theta$ are the current parameters of our classifier) using an older policy $\pi_{\zeta_{\bar{\theta}}}$ (where $\bar{\theta}$ were the parameters of the classifier which induced the constraint set that this policy respects). We can do so by noting that
\begin{equation}
\begin{split}
    Z_\theta
	&= \int\exp(r(\tau)) \zeta_\theta(\tau)d\tau\\
	&= \int \pi_{\zeta_{\bar{\theta}}}(\tau) \left[\frac{\exp(r(\tau)) \zeta_\theta(\tau)}{\pi_{\zeta_{\bar{\theta}}}(\tau)}\right]d\tau\\
	&= \mathbb{E}_{\tau \sim \pi_{\zeta_{\bar{\theta}}}} \left[\frac{\exp(r(\tau)) \zeta_\theta(\tau)}{\pi_{\zeta_{\bar{\theta}}}(\tau)}\right]\\
	&= Z_{\bar{\theta}}\cdot\mathbb{E}_{\tau \sim \pi_{\zeta_{\bar{\theta}}}} \left[\frac{\zeta_\theta(\tau)}{\zeta_{\bar{\theta}}(\tau)}\right].\\
\end{split}
\label{eq:z_theta}
\end{equation}

where the fourth lines follows from our MaxEnt assumption, i.e., $\pi_{\zeta_{\bar{\theta}}}(\tau) = \exp(r(\tau))\zeta_{\bar{\theta}}(\tau)/Z_{\bar{\theta}}.$

Therefore
\begin{equation}
\begin{split}
	\nabla_\theta \log Z_\theta
	&= \frac{1}{Z_\theta} \nabla_\theta Z_\theta\\
	&= \frac{1}{Z_{\bar{\theta}}\cdot\mathbb{E}_{\tau \sim \pi_{\zeta_{\bar{\theta}}}} \left[\frac{\zeta_\theta(\tau)}{\zeta_{\bar{\theta}}(\tau)}\right]}\left[Z_{\bar{\theta}}\cdot\mathbb{E}_{\tau \sim \pi_{\zeta_{\bar{\theta}}}}\left[\frac{\nabla_\theta\zeta_\theta(\tau)}{\zeta_{\bar{\theta}}(\tau)}\right]\right]\\
	&= \frac{1}{\mathbb{E}_{\tau \sim \pi_{\zeta_{\bar{\theta}}}} \left[\frac{\zeta_\theta(\tau)}{\zeta_{\bar{\theta}}(\tau)}\right]}\left[\mathbb{E}_{\tau \sim \pi_{\zeta_{\bar{\theta}}}}\left[ \frac{\zeta_\theta(\tau)}{\zeta_{\bar{\theta}}(\tau)}\nabla_\theta \log \zeta_\theta(\tau)\right]\right].
\end{split}
\end{equation}
Note that $\mathbb{E}_{\tau \sim \pi_{\zeta_{\bar{\theta}}}}\left[\frac{\zeta_\theta(\tau)}{\zeta_{\bar{\theta}}(\tau)}\right] = \int \pi_{\zeta_\theta}(\tau) d\tau =1$. So
\begin{equation}
\begin{split}
    \nabla_\theta \log Z_\theta &= 
    \mathbb{E}_{\pi_{\zeta_{\bar{\theta}}}}\left[\frac{\zeta_\theta(\tau)}{\zeta_{\bar{\theta}}(\tau)}\nabla_\theta \log \zeta_\theta(\tau)\right]\\
    &= \mathbb{E}_{\pi_{\zeta_{\bar{\theta}}}} \left[\prod_{t=1}^T\frac{\zeta_\theta(s_t,a_t)}{\zeta_{\bar{\theta}}(s_t,a_t)	}\nabla_\theta \log \prod_{t'=1}^T\zeta_\theta(s_{t'},a_{t'})\right]\\
    &= \mathbb{E}_{\pi_{\zeta_{\bar{\theta}}}} \left[\prod_{t=1}^T\frac{\zeta_\theta(s_t,a_t)}{\zeta_{\bar{\theta}}(s_t,a_t)	}\sum_{t'=1}^T\nabla_\theta \log \zeta_\theta(s_{t'},a_{t'})\right]\\
    &= \sum_{t'=1}^T\mathbb{E}_{\pi_{\zeta_{\bar{\theta}}}}\left[\prod_{t=1}^T\frac{\zeta_\theta(s_t,a_t)}{\zeta_{\bar{\theta}}(s_t,a_t)	}\nabla_\theta \log \zeta_\theta(s_{t'},a_{t'})\right]\\
    &= \sum_{t'=1}^T\mathbb{E}_{\pi_{\zeta_{\bar{\theta}}}}\left[\left(\prod_{\substack{t=1\\t\neq t'}}^{T}\frac{\zeta_\theta(s_t,a_t)}{\zeta_{\bar{\theta}}(s_t,a_t)}\right)\left(\frac{\zeta_\theta(s_{t'},a_{t'})}{\zeta_{\bar{\theta}}(s_{t'},a_{t'})}\nabla_\theta \log \zeta_\theta(s_{t'},a_{t'})\right)\right]\\
    &= \sum_{t'=1}^T\mathbb{E}_{\tau/(s_{t'},a_{t'})\sim\pi_{\zeta_{\bar{\theta}}}}\left[\prod_{\substack{t=1\\t\neq t'}}^{T}\frac{\zeta_\theta(s_t,a_t)}{\zeta_{\bar{\theta}}(s_t,a_t)}\right]\mathbb{E}_{s_{t'},a_{t'}\sim\pi_{\zeta_{\bar{\theta}}}}\left[\frac{\zeta_\theta(s_{t'},a_{t'})}{\zeta_{\bar{\theta}}(s_{t'},a_{t'})}\nabla_\theta \log \zeta_\theta(s_{t'},a_{t'})\right]\\
    &= \sum_{t'=1}^T \frac{Z_\theta}{Z_{\bar\theta}}\cdot \mathbb{E}_{\pi_{\zeta_{\bar{\theta}}}}\left[\frac{\zeta_\theta(s_{t'},a_{t'})}{\zeta_{\bar{\theta}}(s_{t'},a_{t'})}\nabla_\theta \log \zeta_\theta(s_{t'},a_{t'})\right].\\
    &\approx \sum_{t'=1}^T  \mathbb{E}_{\pi_{\zeta_{\bar{\theta}}}}\left[\frac{\zeta_\theta(s_{t'},a_{t'})}{\zeta_{\bar{\theta}}(s_{t'},a_{t'})}\nabla_\theta \log \zeta_\theta(s_{t'},a_{t'})\right],\\
\end{split}
\end{equation}
where the last step assumes that $Z_\theta \approx Z_{\bar\theta}$. This is justified since we restrict the extent to which $\zeta_\theta$ can change via the early stopping technique.

\subsection{Forward and reverse KL divergences between two policies}
\label{a:kl_is}
Consider two policies $\pi_{\bar\theta}$ and $\pi_{\theta}$. Using our MaxEnt assumption, we can write the forward KL divergence as
\begin{equation}
\begin{split}
    D_{KL}(\pi_{\bar\theta}\vert\vert\pi_\theta)
    &= \mathbb{E}_{\tau\sim\pi_{\bar\theta}} \left[ \log\frac{\pi_{\bar{\theta}}(\tau)}{\pi_\theta(\tau)}\right]\\
    &= \mathbb{E}_{\tau\sim\pi_{\bar\theta}} \left[ \log\frac{\zeta_{\bar{\theta}}(\tau)}{\zeta_\theta(\tau)}\right]+\log\frac{Z_\theta}{Z_{\bar{\theta}}}.
\end{split}
\end{equation}
Let $\omega(\tau)$ denote $\zeta_{\bar{\theta}}(\tau)/\zeta_\theta(\tau)$. Plugging in the expression for $Z_\theta$ from (\ref{eq:z_theta}) and using Jensen's inequality gives
\begin{equation}
\begin{split}
    D_{KL}(\pi_{\bar\theta}\vert\vert\pi_\theta)
    &= \mathbb{E}_{\tau\sim\pi_{\bar\theta}} \left[ \log\omega(\tau)\right] + \log \mathbb{E}_{\tau\sim\pi_{\bar\theta}}\left[\omega(\tau)\right]\\
    &\leq 2\log \mathbb{E}_{\tau\sim\pi_{\bar\theta}}\left[\omega(\tau)\right].
\end{split}
\end{equation}
Similarly, the reverse KL divergence is
\begin{equation}
\begin{split}
    D_{KL}(\pi_{\theta}\vert\vert\pi_{\bar\theta})
    &= \mathbb{E}_{\tau\sim\pi_{\theta}} \left[ \log\frac{\pi_{\theta}(\tau)}{\pi_{\bar\theta}(\tau)}\right]\\
    &= \mathbb{E}_{\tau\sim\pi_{\bar\theta}} \left[\frac{\pi_{\theta}(\tau)}{\pi_{\bar\theta}(\tau)}\log\frac{\pi_{\theta}(\tau)}{\pi_{\bar\theta}(\tau)}\right]\\
    &= \mathbb{E}_{\tau\sim\pi_{\bar\theta}} \left[\omega(\tau)\frac{Z_{\bar\theta}}{Z_\theta}\log\omega(\tau)\frac{Z_{\bar\theta}}{Z_{\theta}}\right]\\
    &= \mathbb{E}_{\tau\sim\pi_{\bar\theta}} \left[\omega(\tau)\log\omega(\tau)\right]\frac{Z_{\bar\theta}}{Z_\theta} + \mathbb{E}_{\tau\sim\pi_{\bar\theta}} \left[\omega(\tau)\right]\frac{Z_{\bar\theta}}{Z_\theta}\log\frac{Z_{\bar\theta}}{Z_{\theta}}.\\
\end{split}
\end{equation}
From (\ref{eq:z_theta}) we know that $Z_{\bar\theta}/Z_\theta=1/\mathbb{E}_{\tau\sim\pi_{\bar\theta}}\omega(\tau)$. Using Jensen's inequality we have
\begin{equation}
\begin{split}
    D_{KL}(\pi_{\theta}\vert\vert\pi_{\bar\theta})
    &= \frac{1}{\mathbb{E}_{\tau\sim\pi_{\bar\theta}}\left[\omega(\tau)\right]}\mathbb{E}_{\tau\sim\pi_{\bar\theta}} \left[\omega(\tau)\log\omega(\tau)\right] - \log \mathbb{E}_{\tau\sim\pi_{\bar\theta}}\left[\omega(\tau)\right]\\
    &\leq \frac{1}{\mathbb{E}_{\tau\sim\pi_{\bar\theta}}\left[\omega(\tau)\right]}\mathbb{E}_{\tau\sim\pi_{\bar\theta}} \left[\omega(\tau)\log\omega(\tau)\right] - \mathbb{E}_{\tau\sim\pi_{\bar\theta}}\left[\log\omega(\tau)\right].\\
\end{split}
\end{equation}
Letting $\bar{\omega}$ denote $\mathbb{E}_{\tau\sim\pi_{\bar\theta}}[\omega(\tau)]$ gives us
\begin{equation}
    D_{KL}(\pi_{\theta}\vert\vert\pi_{\bar\theta}) \leq \frac{\mathbb{E}_{\tau\sim\pi_{\bar\theta}} \left[(\omega(\tau)-\bar\omega)\log\omega(\tau)\right]}{\bar{\omega}}.
\end{equation}

\subsection{Rationale for (\ref{eq:min_max})}
\label{a:maxent}
Consider a constrained MDP $\mathcal{M}^\mathcal{C}$ as defined in Section \ref{sec:crl}. We are interested in recovering the following policy
\begin{equation}
    \pi_{\mathcal{M}^\mathcal{C}}(\tau) = \frac{\exp(\beta r(\tau))}{Z_{\mathcal{M}^\mathcal{C}}}\mathbbm{1}^\mathcal{C}(\tau)
    \label{eq:maxent_cons}
\end{equation}
where $Z_{\mathcal{M}^\mathcal{C}} = \int \exp(\beta r(\tau))\mathbbm{1}^\mathcal{C}(\tau) d\tau$ is the partition function and $\mathbbm{1}^\mathcal{C}$ is an indicator function that is $0$ if $\tau\in\mathcal{C}$ and $1$ otherwise.

\textbf{Lemma:} The Boltzmann policy $\pi^B(\tau) = \exp(\beta r(\tau))/Z$ maximizes $\mathcal{L}(\pi) = \mathbb{E}_{\tau\sim\pi}[r(\tau)] + \frac{1}{\beta}\mathcal{H}(\pi)$, where $H(\pi)$ denotes the entropy of $\pi$.

\textbf{Proof:} Note that the KL-divergence, $D_{KL}$, between a policy $\pi$ and $\pi^B$ can be written as
\begin{equation}
    \begin{split}
    \mathcal{D}_{KL}(\pi\vert\vert\pi^B)
    &= \mathbb{E}_{\tau\sim\pi}[\log\pi(\tau)-\log\pi^B(\tau)]\\
    &= \mathbb{E}_{\tau\sim\pi}[\log\pi(\tau)-\beta r(\tau)+\log Z]\\
    &=-\mathbb{E}_{\tau\sim\pi}[\beta r(\tau)]-\mathcal{H}(\pi) + \log Z\\
    &=-\beta\mathcal{L}(\pi) + \log Z.
    \end{split}
\end{equation}
Since $\log Z$ is constant, minimizing $D_{KL}(\pi\vert\vert\pi^B)$ is equivalent to maximizing $\mathcal{L}(\pi)$. Also, we know that $\mathcal{D}_{KL}(\pi\vert\vert\pi^B)$ is minimized when $\pi = \pi^B$. Therefore, $\pi^B$ maximizes $\mathcal{L}$.

\textbf{Proposition:} The policy in (\ref{eq:maxent_cons}) is a solution of
\begin{equation}
	\underset{\lambda \geq 0}{\text{minimize}} \max_\pi \mathbb{E}_{\tau\sim\pi}[r(\tau)] +\frac{1}{\beta}\mathcal{H}(\pi^{\phi}) - \lambda (\mathbb{E}_{\tau\sim\pi^{\phi}}[\bar{\zeta}_\theta(\tau)] - \alpha).
\end{equation}

\textbf{Proof:} Let us rewrite the inner optimization problem as
\begin{equation}
    \max_\pi\; \mathbb{E}_{\tau\sim\pi}[r(\tau)-\lambda (\bar{\zeta}_\theta(\tau)-\alpha)] + \frac{1}{\beta}\mathcal{H}(\pi).
\end{equation}
From the Lemma we know that the solution to this is
\begin{equation}
    \pi(\tau,\lambda) = \frac{g(\tau,\lambda)}{\int g(\tau',\lambda) d\tau'},
\end{equation}
where $g(\tau,\lambda)=\exp(\beta (r(\tau)-\lambda (\bar{\zeta}_\theta(\tau)-\alpha)))$. To find $\pi^*(\tau) = \min_\lambda \pi(\tau,\lambda)$, note that:
\begin{enumerate}
    \item When $\bar{\zeta}_\theta(\tau) \leq \alpha$, then $\lambda^* = 0$ minimizes $\pi$. In this case $g(\tau,\lambda^*) = \exp(\beta r(\tau))$.
    \item When $\bar{\zeta}_\theta(\tau)>\alpha$, then $\lambda^* \rightarrow \infty$ minimizes $\pi$. In this case $g(\tau,\lambda^*)=0$.
\end{enumerate}
We can combine both of these cases by writing
\begin{equation}
    \pi^*(\tau) = \frac{\exp(r(\tau))}{\int \exp(r(\tau'))\mathbbm{1}^{\bar{\zeta}_\theta}(\tau')d\tau'}\mathbbm{1}^{\bar{\zeta}_\theta}(\tau),
\end{equation}
where $\mathbbm{1}^{\bar{\zeta}_\theta}(\tau)$ is $1$ if $\bar{\zeta}_\theta(\tau) \leq \alpha$ and $0$ otherwise. (Note that the denominator is greater than $0$ as long as we have at least one $\tau$ for which $\bar{\zeta}_\theta(\tau) \leq \alpha$, i.e., we have at least one feasible solution.)

\subsection{Experimental settings}
\label{a:es}
We used W\&B \citep{wandb2020} to manage our experiments and conduct sweeps on hyperparameters. We used Adam \citep{kingma2015adam} to optimize all of our networks. All important hyperparameters are listed in Table \ref{table:hp}. For the ablation studies we used the same parameters as listed in the table for HalfCheetah. Details on the environments can be found below.

\subsubsection{LapGridWorld}
Here, agents move on a $11 \times 11$ grid by taking either clockwise or anti-clockwise actions. The agent is awarded a reward $3$ each time it moves onto a bridge with a dollar (see Figure \ref{fig:envs}). The agent's state is the number of the grid it is on.

\subsubsection{HalfCheetah, Ant and Ant-Broken}
The original reward schemes for HalfCheetah and Ant in OpenAI Gym \citep{brockman2016openai}, reward the agents proportional to the distance they cover in the forward direction. We modify this and instead simply reward the agents according to the amount of distance they cover (irrespective of the direction they move in). For Ant-Broken we simply disable two of the legs of Ant by hard-coding a torque of $0$ on their motors.

\subsubsection{Point}
For the Point agent, the reward function at each timestep is defined as follows
\begin{equation}
    r := \frac{ydx - xdx}{\left(1 + \vert \sqrt{x^2+y^2}-10)\vert\right)}
\end{equation}
where $x, y$ are the position coordinates of the agent and $dx$ and $dy$ are the distances that the agent has moved in $x$ and $y$ directions respectively in that timestep.

\begin{table}[th]
\caption{List of hyperparameters. For neural network architectures we report the number of hidden units in each layer. All hidden layers use the tanh activation function.}
\label{table:hp}
\vskip 0.15in
\begin{center}
\begin{small}
\begin{sc}
\begin{tabular}{lcccccr}
\toprule
Parameter & LapGridWorld & HalfCheetah & Ant & Point & AntBroken\\
\midrule
Policy, $\pi_\phi$\\
\;\;\;\;Architecture       & \\
\;\;\;\;\;\;\;Policy Network & 64, 64 & 64, 64 &64, 64& 64, 64 & 64, 64 \\
\;\;\;\;\;\;\;Value Network & 64, 64 & 64, 64 &64, 64& 64, 64 & 64, 64 \\
\;\;\;\;\;\;\;Cost Value Network & 64, 64 & 64, 64 &64, 64& 64, 64 & 64, 64 \\
\;\;\;\;Batch Size      & 64   & 64   & 128  & 64   & 128\\
\;\;\;\;PPO Target KL   & $0.01$ & 0.01 & 0.01 & 0.01 & 0.01\\
\;\;\;\;Learning Rate   & $3 \times 10^{-4}$  & $3 \times 10^{-4}$  & $3 \times 10^{-5}$  & $3 \times 10^{-4}$  & $3 \times 10^{-5}$ \\
\;\;\;\;Reward-GAE-$\gamma$      & 0.99 & 0.99 & 0.99 & 0.99 & 0.99\\
\;\;\;\;Reward-GAE-$\lambda$     & 0.95 & 0.95 & 0.90 & 0.95 & 0.90\\
\;\;\;\;Cost-GAE-$\gamma$        & 0.99 & 0.99 & 0.99 & 0.99 & 0.99\\
\;\;\;\;Cost-GAE-$\lambda$       & 0.95 & 0.95 & 0.95 & 0.95 & 0.95\\
\;\;\;\;Entropy Bonus, $1/\beta$ & 0.0  & 0.0  & 0.0  & 0.0  & 0.0\\
Lagrangian, $\lambda$\\
\;\;\;\;Initial Value  & 1.0 & 1.0 & 0.1 & 1.0 & 0.1\\
\;\;\;\;Learning Rate  & 0.1 & 0.1 & 1.0 & 0.1 & 1.0\\
\;\;\;\;Budget         & 0.0 & 0.0 & 0.0 & 0.0 & 0.0\\
Constraint Function, $\zeta_\theta$\\
\;\;\;\;Architecture                    & 20   & 20   & 40,40 & - & -\\
\;\;\;\;Learning Rate                   & 0.01 & 0.01 & 0.01  & - & -\\
\;\;\;\;Backward Iterations             & 10   & 10   & 10    & - & -\\
\;\;\;\;Regularizer Weight              & 0.5  & 0.5  & 0.6   & - & -\\
\;\;\;\;Max Forward KL, $\epsilon_F$    & 10   & 10   & 10    & - & -\\
\;\;\;\;Max Backward KL, $\epsilon_B$   & 2.5  & 2.5  & 2.5   & - & -\\
Miscellaneous\\
\;\;\;\;Expert Rollouts                 & 1    & 10   & 45    & -   & -\\
\;\;\;\;Rollout Length                  & 200  & 1000 & 500   & 150 & 500\\
\bottomrule
\end{tabular}
\end{sc}
\end{small}
\end{center}
\vskip -0.1in
\end{table}

\end{document}